\documentclass[11pt, a4paper, logo]{lumia}

\usepackage[numbers,sort&compress]{natbib}

\usepackage[utf8]{inputenc}
\usepackage[T1]{fontenc}
\usepackage{url}
\usepackage{booktabs}
\usepackage{amsfonts}
\usepackage{amsmath}
\usepackage{nicefrac}
\usepackage{microtype}
\usepackage{xcolor}
\usepackage{graphicx}
\usepackage{multirow}
\usepackage{makecell}
\usepackage{colortbl}
\usepackage{tabularx}
\usepackage{enumitem}
\usepackage{hyperref}
\usepackage{adjustbox}

\definecolor{abstrabg}{HTML}{F2D1C5}
\definecolor{abstradark}{HTML}{D97757}
\definecolor{darkblue}{rgb}{0, 0, 0.5}
\definecolor{lightblue}{HTML}{D0CFEA}

\usepackage{xspace}

\theoremstyle{plain}

\newtheorem*{proposition*}{Proposition}

\theoremstyle{definition}

\theoremstyle{definition}

\def\eqref#1{equation~\ref{#1}}

\usepackage{float}
\usepackage{wrapfig}

\usepackage{tcolorbox}

\definecolor{highval}{HTML}{E8F5E9}
\definecolor{medval}{HTML}{FFF9C4}
\definecolor{lowval}{HTML}{FFEBEE}

\usetikzlibrary{calc,positioning,shapes,arrows.meta,fit,backgrounds}
\graphicspath{{figures/}}

\definecolor{citecolor}{HTML}{0071bc}

\hypersetup{
    colorlinks=true,
    linkcolor=red,
    citecolor=citecolor,
    filecolor=magenta,      
    urlcolor=magenta,
}

\correspondingemail{\emailicon~\href{jiasheng.tjs@alibaba-inc.com}{jiasheng.tjs@alibaba-inc.com}; \emailicon~\href{wangbo.zhao96@gmail.com}{wangbo.zhao96@gmail.com}; \emailicon~\href{yangyou@nus.edu.sg}{yangyou@nus.edu.sg} \quad $^\ddagger$ Corresponding Author\\
Homepage: \href{https://www.omnisource.cn/agent-as-a-router}{https://omnisource.cn/agent-as-a-router}
}

\title{Agent-as-a-Router: Agentic Model Routing for Coding Tasks}

\author{
\parbox{0.85\textwidth}{\centering
  Pengfei Zhou$^{1,2}$\quad Zhiwei Tang$^{2,3,4}$\quad Yixing Ma$^{5}$\quad
  Jiasheng Tang$^{2,3}$$^\ddagger$\quad Yizeng Han$^{2}$\quad Zhenglin Wan$^{1}$\quad
  Fanqing Meng$^{1}$\quad Wei Wang$^{6}$\quad Bohan Zhuang$^{4}$\quad
  Wangbo Zhao$^{6}$$^\ddagger$\quad Yang You$^{1}$$^\ddagger$\\
  $^{1}$National University of Singapore \quad
  $^{2}$DAMO Academy, Alibaba Group \quad
  $^{3}$Hupan Lab \quad
  $^{4}$Zhejiang University\\
  $^{5}$University of California, Berkeley \quad
  $^{6}$The Hong Kong University of Science and Technology 
}
}

\begin{document}



\begin{abstract}
Real-world users typically have access to multiple Large Language Models (LLMs) from different providers, and these LLMs often excel at distinct domains, yet none dominate all. Consequently, routing each task to the most suitable model becomes critical for both performance and cost. Existing routers treat this as a static, one-off classification problem. However, we identify the performance bottleneck for these routers as \emph{information deficit}: simply augmenting a vanilla LLM router with performance statistics at the task-dimension level yields a 15.3\% relative gain, surpassing a heuristic router built on the same dimension-level priors. Motivated by this finding, we propose \textbf{Agent-as-a-Router}, a framework that formalizes routing as a C-A-F loop (Context$\to$Action$\to$Feedback$\to$Context). It closes the information gap by accumulating execution-grounded experience during deployment. We instantiate this framework as \textbf{ACRouter}, composed of an Orchestrator, a Verifier, a Memory module, and introduce \textbf{CodeRouterBench}, an evaluation environment comprising $\sim$10K task instances with verified scores from 8 frontier LLMs, enabling regret-based router comparison on streaming tasks. Experiments show that ACRouter achieves the lowest cumulative regret on in-distribution tasks and generalizes to out-of-distribution agentic-programming tasks, demonstrating that our routing framework actively closes the information gap. Codes and benchmarks are released at \href{https://github.com/LanceZPF/agent-as-a-router}{https://github.com/LanceZPF/agent-as-a-router}.
\end{abstract}

\maketitle
\section{Introduction}
\label{sec:intro}
Modern coding agents such as Claude Code~\citep{anthropic2025claudecode} and Codex~\citep{openai2025codex} have had a significant impact on real-world software development by turning LLMs into interactive systems for coding, debugging, and repository-level programming. However, most of these agents tend to solve all tasks using the same Large Language Model (LLM)~\citep{yue2025masrouter}. While this design is reasonable from a \emph{provider-centric} serving perspective, where providers prioritize in-house models and predictable serving costs~\citep{research2026composer}, it overlooks the actual needs of users in \emph{user-centric} scenarios, where the priority is task-level quality and cost-efficiency rather than provider-side predictability.

In such scenarios, users can subscribe to multiple providers and run capable open-source models locally. Across our experiments of 8 frontier models on various coding dimensions (Fig.~\ref{fig:complementarity}), the best model varies per task, and always picking the globally strongest model still lags behind the per-task oracle (chooses the best model for each task). As manually selecting the best model for each task is infeasible at scale, a critical question emerges: \emph{which model should handle each incoming task?} This motivates automatic model routing as a key mechanism for improving agent performance.

Existing routing methods typically frame this as a static classification problem, employing language models as the routing policy~\citep{ong2024routellm, liu2026adaptive, varshney2026llm}. However, our preliminary experiments reveal that a zero-shot LLM-as-a-Router, even when powered by a highly capable model like Claude Sonnet 4.6, still falls short of the per-task oracle by a wide margin (see Table~\ref{tab:bias}). This substantial performance gap suggests that the fundamental bottleneck in model routing extends beyond pure reasoning capabilities.

\textbf{What is actually limiting these routers? Reasoning capability, or information access?} To find out, we run an ablation that varies \emph{only} the information available to the LLM router (Table~\ref{tab:bias}). With only the zero-shot prompt Vanilla router scores 41.41, and adding per-dimension performance statistics from a held-out probing set (+Perf stats) improves the score to 47.74, a +15.3\% relative gain over Vanilla. It also \emph{exceeds} the best heuristic (scoring 47.50) that encodes the same dimension-level statistics information. Therefore, we find that the bottleneck for model routing is \emph{information deficit} rather than \emph{reasoning failure} (\S\ref{sec:diagnosis}).


To close this performance gap, a router must \emph{acquire and accumulate} execution-grounded information during deployment. Static routers are structurally unable to do this since their information state is frozen. This motivates a different class of self-adaptive router, one that evolves over the task stream, verifies each decision, and conditions future decisions on accumulated expertise.

\begin{figure*}[!t]
\centering
\includegraphics[width=\textwidth]{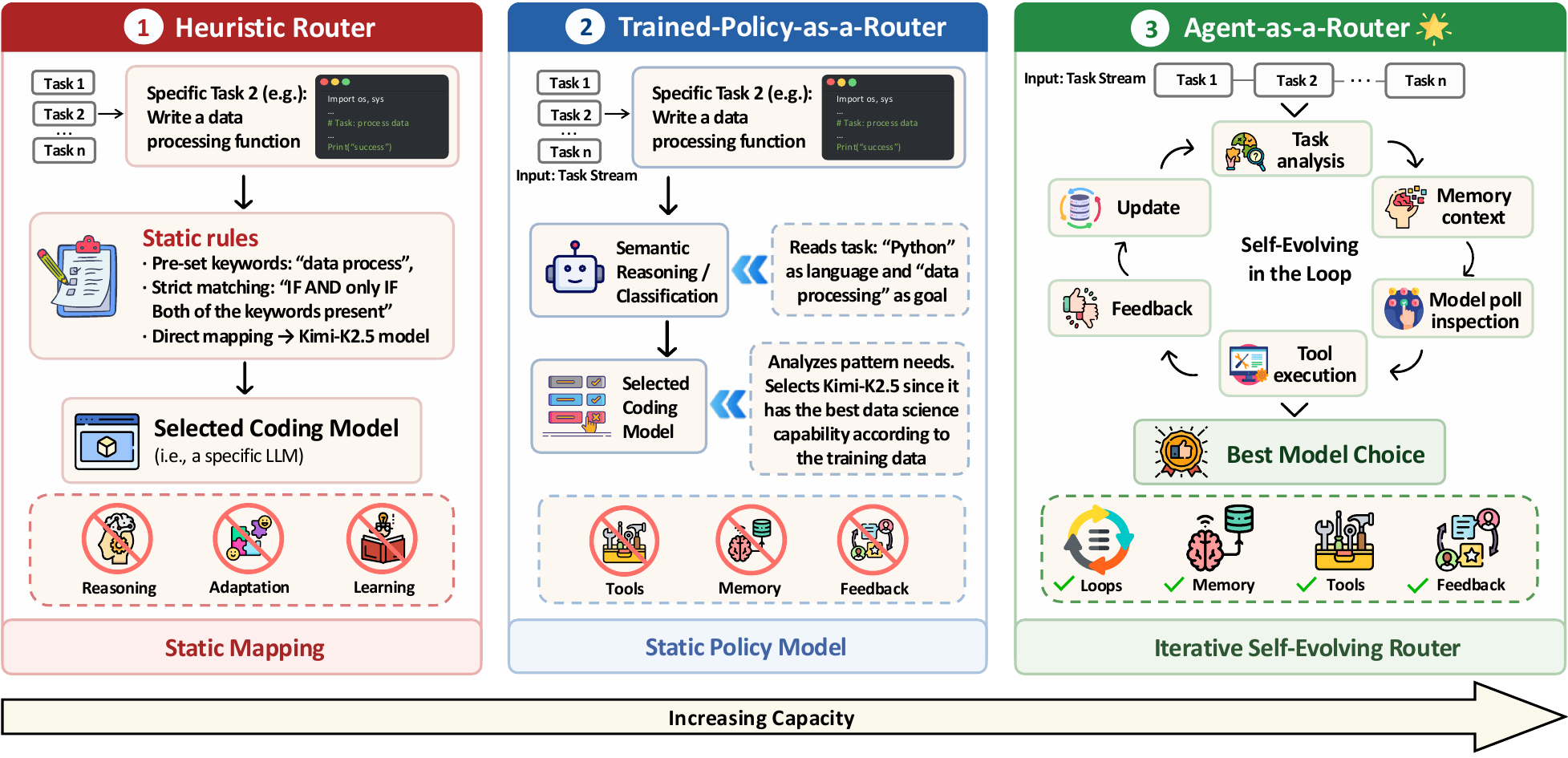}
\caption{\textbf{Comparison of three routing strategies.} \textbf{(1)} Static, heuristic-based routers (router directly dispatches via a lookup table, e.g.\ DimensionBest). \textbf{(2)} Routers based on a static trained policy (router uses a learned policy model with no memory). \textbf{(3)} Our proposed Agent-as-a-Router (router with iterative self-evolving capabilities in the task stream).}
\label{fig:teaser_compare}
\end{figure*}

We propose \textbf{Agent-as-a-Router} framework, formalizing routing as a Context-Action-Feedback (C-A-F) loop, in which each loop's verified outcome enters the next loop's context (Fig.~\ref{fig:teaser_compare}). The router observes a \emph{Context} (prior plus accumulated experience), selects an \emph{Action} (which model to invoke), receives verification \emph{Feedback} (score and cost-efficiency), and merges the feedback back into the \emph{Context} for the next task. This loop relates to a \emph{contextual multi-armed bandit}~\citep{li2010contextual}, so \emph{cumulative regret} (the running gap to the per-task oracle) becomes the natural streaming metric. 

We instantiate the framework as \textbf{ACRouter} (\textbf{A}gentic \textbf{C}oding \textbf{Router}), comprising three core modules (Orchestrator, Verifier, and Memory) and backed by a comprehensive toolkit that includes diverse routing policies, sandboxed execution environments and validation tools.
We evaluate ACRouter on our established \textbf{CodeRouterBench}, which contains $\sim$10K tasks across 9 in-distribution (ID) coding dimensions and an out-of-distribution (OOD) agentic-programming testbed with verified scores from 8 frontier LLMs to enable regret-based comparison on streaming tasks~\citep{openai_swebench_verified,feng2026longcli,chen2026swe,zhou2026featurebench}. ACRouter attains the lowest cumulative regret across all evaluated routers on ID streams and also generalizes well to OOD tasks, consistently outperforming other routers.

Our contributions are threefold:
(1) \textbf{Framework.} We propose \textbf{Agent-as-a-Router}, formalizing model routing as a Context-Action-Feedback (C-A-F) loop, with cumulative regret as the natural streaming metric.
(2) \textbf{Artifacts.} We build \textbf{ACRouter} as a C-A-F instantiation, and present \textbf{CodeRouterBench} ($\sim$10K tasks, 8 LLMs, execution-verified) for regret-based router evaluation.
(3) \textbf{Findings.} \emph{Information deficit} rather than reasoning is the routing bottleneck (+15.3\% when given per-dimension performance statistics); ACRouter attains the lowest cumulative regret on both in-distribution and OOD tasks, while lightweight static routers fail to generalize on OOD tasks.

\section{Related Work}
\label{sec:related}

\paragraph{LLM Routing.}
The problem of selecting among multiple LLMs for a given query has attracted growing attention~\citep{dong2024automix,chen2023frugalgpt,ding2024hybrid}. 
RouteLLM~\citep{ong2024routellm} formulated routing as a preference learning problem, training classifiers on human preference data to predict which of two models produces better responses. Meta-modeling approaches~\citep{vsakota2024fly, shnitzer2023largelanguagemodel} learn to predict model performance from task features. Most recently, LLMRouterBench~\citep{duwal2026llmrouterbench} evaluated routing across 21 general NLU datasets with 33 models. Our work differs from these by proposing \textbf{Agent-as-a-Router} and formalizing it as the C-A-F loop for adaptive routing. Moreover, we specifically benchmark routers in an agentic coding setting.

\paragraph{Coding Agent.}
Coding agents have evolved from single-call code generators~\citep{austin2021mbpp, chen2021humaneval} into multi-stage harness-based frameworks that interleave planning, retrieval, code editing, execution, and self-debugging on repository-level tasks~\citep{jimenez2024swebench, yang2024sweagent, xia2024agentless, wang2024openhands}. Multi-agent variants further decompose these stages across specialized roles~\citep{hong2024metagpt, qian2024chatdev}. Production systems further integrate these features into deployed assistants~\citep{anthropic2025claudecode, openai2025codex}.
However, existing frameworks typically rely on a fixed LLM backbone, rather than dynamically selecting the best model for each specific task.
\textbf{ACRouter} addresses this limitation by actively routing each task to the most suitable model within a continuous stream. To support the evaluation of this framework, \textbf{CodeRouterBench} provides a standardized streaming environment to compare different routing methods using cumulative regret.

\section{Agent-as-a-Router}
\label{sec:agentmethod}

\subsection{Preliminary: The Performance Gap Diagnosis}
\label{sec:diagnosis}

\begin{table}[h]
\caption{Preliminary ablation study diagnosing the performance bottleneck in LLM-as-a-Router (Claude Sonnet 4.6 tested on 2{,}919 tasks). AvgPerf: average performance score. Perf/\$: AvgPerf\% per USD. Providing prior performance statistics significantly improves routing performance.}
\vspace{-0.5em}
\label{tab:bias}
\centering
\small
\begin{tabular}{llcc}
\toprule
\textbf{Ablation} & \textbf{Interpretation} & \textbf{AvgPerf\%} & \textbf{Perf/\$} \\
\midrule
Oracle & Theoretical upper bound using the best model for each task & \textbf{57.00} & 8.20 \\ 
DimensionBest & Select the best model for each dimension by prior & 47.50 & 3.69  \\
\midrule
Vanilla & Standard zero-shot LLM-as-a-router & 41.41 & 1.97  \\
+Dimension & +Task dimension description & 41.18 & 1.81 \\
\textbf{+Perf stats} & +Prior performance statistics from a probing set & \textbf{47.74} & 1.71  \\
\bottomrule
\end{tabular}
\end{table}

We first conduct a preliminary experiment to diagnose the performance bottleneck of LLM-as-a-Router (Table~\ref{tab:bias}). The Vanilla baseline is a standard LLM router (all using Claude Sonnet 4.6) that selects a model from the candidate pool given only the zero-shot task prompt. 
\emph{+Dimension} additionally reveals the task's coding dimension information, and \emph{+Perf stats} further exposes per-dimension performance statistics collected on a separate probing set (7,080 tasks). We compare these variants against DimensionBest, which selects the best model for each dimension with full priors.

When given the same statistics that DimensionBest encodes, the LLM router exceeds it ($47.74$ vs.\ $47.50$ AvgPerf\%) and improves over the Vanilla baseline by a relative $15.3\%$ (from $41.41$ to $47.74$ AvgPerf\%). 
This suggests that \textbf{a major source of the performance gap between LLM-as-a-Router and the oracle upper bound is information deficit, rather than a lack of reasoning capability.}

Two design insights follow from this diagnosis: (i) the router must \emph{acquire} new execution-grounded information at each decision—that is, performance signals generated by actually running the selected model's output in a sandbox rather than relying on static priors or model self-assessment (verification); and (ii) the router must \emph{accumulate} it across the task stream so that future decisions can condition on past outcomes (memory). We formalize these insights through the C-A-F loop (\S\ref{sec:framework}) and instantiate it as ACRouter (\S\ref{sec:acrouter}).


\subsection{The C-A-F Loop}
\label{sec:framework}

Following the diagnosis in \S\ref{sec:diagnosis}, we now formalize \textbf{Agent-as-a-Router} that operates over the task stream and updates its internal state from each loop's verified outcome. Concretely, the router has access to an indexed model pool $\mathcal{M}=\{m_1,\dots,m_M\}$ with $M$ models and processes a stream of $N$ tasks $\mathcal{T}=(t_1,\dots,t_N)$ one by one. After each routing decision, the verified outcome is fed back into the context for the next decision, yielding the Context-Action-Feedback (C-A-F) loop below.

\paragraph{The C-A-F loop.}
At task $t_i$, the router observes Context $c_i$, selects Action $a_i\in[M]$, and receives verification Feedback $f_i$, which is memorized into $c_{i+1}$:
\begin{equation}
c_i \;\xrightarrow{\;\text{Decide}\;}\; a_i \;\xrightarrow{\;\text{Execute}\;}\; f_i \;\xrightarrow{\;\text{Memorize}\;}\; c_{i+1}.
\label{eq:cap_cycle}
\end{equation}
We refer to this as the C-A-F loop (Context, Action, Feedback), where each completed loop makes the next one more informed. The loop C$\to$A$\to$F$\to$C repeats as the task stream advances.

\paragraph{Per-loop components.} \textbf{Context} $c_i = (p_i,\; d_i,\; \mathcal{H}_{<i})$: $p_i$ is the input prompt of task $t_i$, $d_i$ is optional metadata (description, difficulty, language), and $\mathcal{H}_{<i}$ denotes the Memory state accumulated from all prior loops; \textbf{Action} $a_i \in [M]$: the index of the selected model $m_{a_i}$ from the model pool $\mathcal{M}$; \textbf{Feedback} $f_i = (\hat{s}_i,\; \hat{\kappa}_i)$: $\hat{s}_i \in [0,1]$ is the verifier-observed score for the selected model $m_{a_i}$, and $\hat{\kappa}_i$ is the corresponding monetary cost computed from token consumption and official prices. This constitutes the execution-grounded feedback that the Memory module accumulates.

\paragraph{Contextual-bandit equivalence.}
The C-A-F formulation relates to a contextual multi-armed bandit problem~\citep{li2010contextual, lattimore2020bandit} with $c_i$ as side information, $a_i$ as the arm pull, and $f_i$ as the feedback. 
Define the history at task $t_i$ as $h_i = (c_1, a_1, f_1,\, \ldots,\, c_{i-1}, a_{i-1}, f_{i-1},\, c_i)$; a routing policy induces a multinomial distribution over the model pool, i.e., $\pi(\cdot \mid h_i) \in \Delta^{M-1}$, where $\Delta^{M-1}$ denotes the probability simplex over $M$ actions. 
Per-task reward combines performance and cost under user-specified weights $\epsilon_1,\epsilon_2\in\mathbb{R}$ ($\epsilon_1>0$ to reward performance, $\epsilon_2<0$ to penalize cost):
\begin{equation}
  r_i(a_i) \;=\; \epsilon_1\,s_i(a_i) + \epsilon_2\,\kappa_i(a_i),
\label{eq:reward1}
\end{equation}
where $s_i(a_i)$ and $\kappa_i(a_i)$ denote the ground-truth score and cost of the selected model on task $t_i$, respectively. The policy's mean reward over the stream is
\begin{equation}
  V(\pi) \;=\; \frac{1}{N}\sum_{i=1}^{N} r_i(a_i)
  \;=\;
  \epsilon_1\frac{1}{N}\sum_{i=1}^{N}s_i(a_i)
  +
  \epsilon_2\frac{1}{N}\sum_{i=1}^{N}\kappa_i(a_i) .
\label{eq:reward_mean}
\end{equation}

\paragraph{Per-task oracle and cumulative regret.}
To compare routers under identical conditions, we pre-construct a full outcome matrix $O\in\mathbb{R}^{N\times M\times 2}$, where $O_{ij} = (s_{ij}, \kappa_{ij})$ stores the ground-truth score and cost of model $m_j$ on task $t_i$. The induced reward matrix $R \in \mathbb{R}^{N \times M}$ is
\begin{equation}
R_{ij} \;=\; \epsilon_1\,s_{ij} + \epsilon_2\,\kappa_{ij}
\quad\text{for } i\in[N],\,j\in[M].
\label{eq:reward_matrix}
\end{equation}
The per-task oracle independently selects the reward-maximizing model for each task with full prior knowledge of $R$:
\begin{equation}
a_i^* = \arg\max_{j \in [M]} R_{ij}, \qquad  r_i^* \;=\; \max_{j \in [M]} R_{ij}, \quad \forall i = 1, \dots, N,
\label{eq:oracle}
\end{equation}
so the oracle's overall mean reward is
\begin{equation}
V^* = \frac{1}{N} \sum_{i=1}^{N} r_i^* = \frac{1}{N} \sum_{i=1}^{N} \max_{j \in [M]} R_{ij}.
\label{eq:oracle_perf}
\end{equation}
Note that this per-task oracle is generally not equal to a single-best-arm policy that commits to one global optimal model. Given a policy $\pi$, we report \textbf{cumulative regret}:
\begin{equation}
\mathrm{CumReg}_N(\pi) \;=\; \sum_{i=1}^{N}\delta_i
  \;=\; N\bigl(V^* - V(\pi)\bigr),
\label{eq:regret}
\end{equation}
where $\delta_i = r_i^* - r_i(a_i) \ge 0$ is the single-task regret. Cumulative regret measures the accumulated reward gap over the task stream, with lower values indicating routing closer to optimal.

\subsection{ACRouter Instantiation}
\label{sec:acrouter}

\begin{figure*}[t]
\centering
\includegraphics[width=\textwidth]{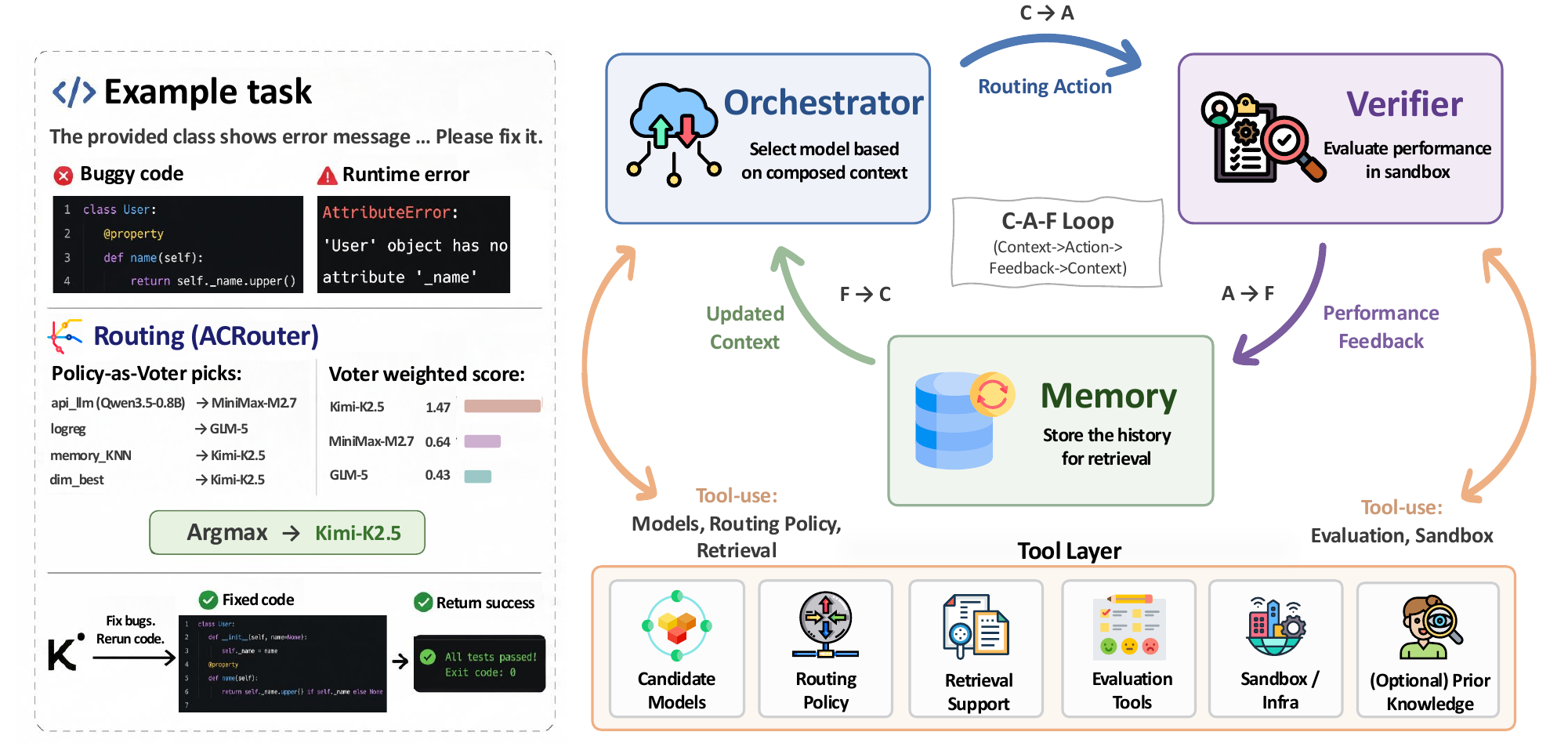}
\vspace{-1.5em}
\caption{\textbf{Architecture of our proposed ACRouter.} It instantiates the C-A-F formulation through a continuous loop: an Orchestrator ensembles multiple signals to make routing decisions, a Verifier evaluates execution in sandbox, and a Memory module stores the feedback to improve future routing.}
\vspace{-1em}
\label{fig:agent_design}
\end{figure*}

The diagnosis in \S\ref{sec:diagnosis} indicates three needs: (i) integrate available information at decision time, (ii) produce new execution-grounded information at each loop, and (iii) accumulate it across loops so that future decisions condition on past outcomes. \textbf{ACRouter} (Fig.~\ref{fig:agent_design}) realizes these as Orchestrator, Verifier, and Memory, respectively, and evolves over the task stream with all three modules active.

\textbf{Orchestrator (integrating information).} The Orchestrator makes the routing decision based on dynamic context: the DimensionBest prior, the top-$10$ historical neighbors retrieved from Memory by kNN, and task metadata. The core policy is a cost-effective Qwen3.5-0.8B model fine-tuned on the CodeRouterBench probing set, combined with heuristic rules via weighted voting.

\textbf{Verifier (producing information).} The Verifier evaluates model output and aggregates multiple signals into a unified performance score $u_i \in [0,1]$ for the current task $t_i$:
\begin{equation}
u_i = \sum_{k \in \mathcal{K}_{d(t_i)}} w_{d(t_i), k} \cdot \hat{s}_k(a_i, t_i)\ ,
\label{eq:verifier}
\end{equation}
where $d(t_i)$ is the type of task $t_i$ (which determines whether it is executable), $\mathcal{K}_{d(t_i)}$ is the set of verification tools (e.g., AST parsing and sandbox execution), $s_k(\cdot) \in [0,1]$ is the scalar score from the $k$-th tool, and $w_{d(t_i), k}$ are type-specific weights where $\sum_{k \in \mathcal{K}_{d(t_i)}} w_{d(t_i),k}=1$.
The tool layer that supports Orchestrator and Verifier is shown in Fig.~\ref{fig:agent_design} and detailed in Appendix~\ref{sec:agent_arch}.


\textbf{Memory (accumulating information).} Memory is an online vector store keyed by task embeddings (voyage-code-3 / BGE-large) whose value logs the chosen model, performance, cost, and verification traces. During retrieval, it uses cosine kNN to fetch the top-$10$ neighbors, which are then fed to the Orchestrator. The store is FIFO-bounded at $E$ entries ($E$ is set to 20K in our implementation) and is committed in place after each attempt. Unlike static dimension-hashed routing, this embedding-based store enables fine-grained, \emph{context-aware} decision making and makes both past successes and recent failures of any candidate on similar tasks visible to the Orchestrator.

\subsection{Decomposed Routing Policies}
\label{sec:variants}

The C-A-F loop provides a unified perspective to inspect existing routing strategies. By respectively restricting or removing specific components (Orchestrator, Verifier, Memory) from the full framework, we organize several baseline routing policies, which naturally set up the ablation study in \S\ref{sec:experiments}. 



\textbf{Single-Model (no Orchestrator, no Memory, no Verifier).} \emph{Always-$m$} routes every task to a fixed model $m$ regardless of context. Included as a reference performance floor.

\textbf{Static: Heuristic (frozen Memory; no Orchestrator policy, no Verifier).} Hardcoded rules that select models from a frozen prior Memory built from probing-set statistics. \emph{DimensionBest} is a coarse-grained oracle that routes each task to the dimensionally best model using dimension-level prior knowledge. \emph{kNN Retrieval} selects a model based on the task-model pair retrieved from nearest-neighbor tasks in the frozen probing-set memory.

\textbf{Static: Trained Policy (trained Orchestrator; no Memory, no Verifier).} A learned classifier that maps task features directly to a model choice. We evaluate lightweight discriminative routers---LogReg (TF-IDF features), TF-IDF$+$MLP, and RouteLLM~\citep{ong2024routellm} (Matrix Factorization and BERT versions)---all trained on the probing set. We additionally finetune Qwen3.5~\cite{qwen35_blog} variants with Low-Rank Adaptation (LoRA)~\cite{hulora} on the same probing set; the scaling sweep is provided in Appendix~\ref{app:scaling}, and Qwen3.5-0.8B is reported in the main table for fair comparison.

\textbf{Dynamic: Online Bandit (parametric Memory; argmax Orchestrator; reward-only Verifier).} Classical contextual bandits cast routing as a per-task arm-pull, with reward $r_i(a_i) = \epsilon_1\,\mathrm{s}_i(a_i) + \epsilon_2\,\mathrm{\kappa}_i(a_i)$ ($\epsilon_1{=}1, \epsilon_2{=}{-}0.1$ based on only ground-truth without traces). Per-arm parametric posteriors replace ACRouter's cosine-kNN Memory, and the Orchestrator collapses to a single $\arg\max$ rule. We evaluate two disjoint per-arm linear contextual policies---\emph{LinUCB}~\citep{li2010contextual} ($\alpha{=}\lambda{=}1$) and \emph{LinTS}~\citep{agrawal2013thompson} ($v{=}0.5,\lambda{=}1$)---each fed either a categorical (dimension/difficulty/language one-hot) or a $64$-dim Johnson--Lindenstrauss projection of the task embedding as context. Bandits are warm-started on the probing set in a deterministic shuffle (seed 42) and updated online during testing.


\section{CodeRouterBench}
\label{sec:benchmark}

Evaluating cumulative regret on streaming routing requires a controlled environment with pre-collected per-task per-model outcomes; existing routing benchmarks only measure single-shot accuracy and cannot support this evaluation. We therefore introduce \textbf{CodeRouterBench}, a unified testbed consisting of $\sim$10K coding tasks across 10 dimensions (see Table~\ref{tab:dimensions}). CodeRouterBench is designed to be extensible with custom dimensions under the same C-A-F formulation.

\begin{figure*}[t]
\centering
\includegraphics[width=0.98\textwidth]{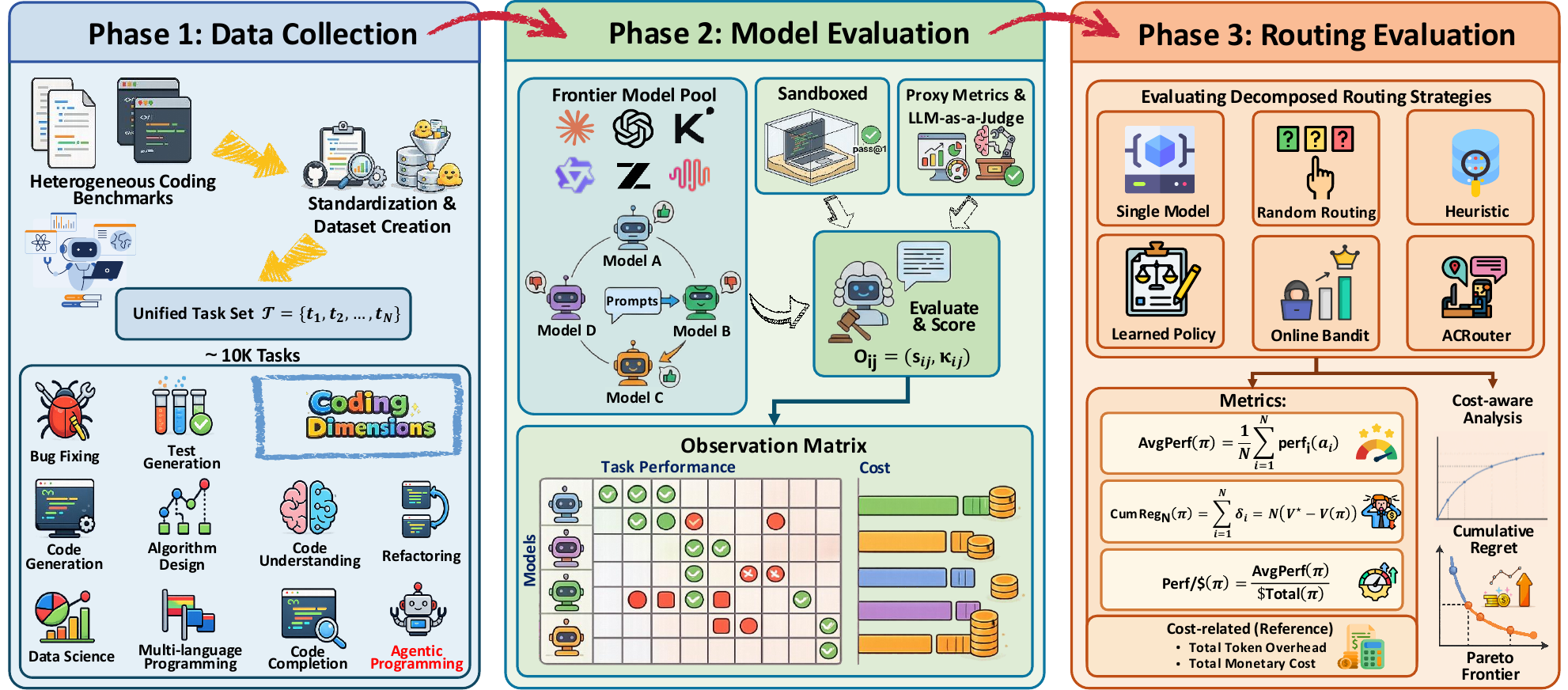}
\vspace{-0.8em}
\caption{\textbf{CodeRouterBench construction and evaluation pipeline.} \textbf{Phase 1:} 15 benchmark sources are unified into $\sim$ 10K tasks across 10 coding dimensions. \textbf{Phase 2:} 8 frontier LLMs generate observation matrix, scored via sandboxed execution and LLM-as-Judge. \textbf{Phase 3:} Several routing methods are evaluated on performance and cost metrics with Pareto analysis.}
\vspace{-1.8em}
\label{fig:framework}
\end{figure*}

\subsection{Benchmark Construction}
\label{sec:construction}

\begin{wraptable}{r}{0.38\linewidth}
\vspace{-50pt}
\caption{CodeRouterBench statistics.}
\vspace{1pt}
\label{tab:dataset_stats}
\centering
\small
\begin{tabular}{lr}
\toprule
\textbf{Statistic} & \textbf{Value} \\
\midrule
Coding dimensions & 10 \\
Source benchmarks & 15+ \\
\midrule
Probing Set Tasks & 7{,}080 \\
In-Distribution Test Tasks &  2{,}919 \\
\midrule
OOD Test Tasks & 176 \\
\bottomrule
\end{tabular}
\vspace{-8pt}
\end{wraptable}
As shown in Fig.~\ref{fig:framework}, all tasks are repurposed from widely-used, high-quality benchmarks~\citep{chen2021humaneval, austin2021mbpp, zhuo2024bigcodebench, jain2024livecodebench, lai2023ds1000, liu2024cruxeval, jimenez2024swebench, zheng2024humaneval_x, cassano2023multipl, tufano2019bugs2fix, cassano2024canitedit, lu2021codexglue}, unifying the evaluation protocol and environment into one framework. Seven dimensions use execution-based scoring (pass@1 in sandboxed environments) for evaluations, while three dimensions use proxy metrics supplemented by LLM-as-Judge. We divide 10,111 tasks into probing and two test sets (see Table~\ref{tab:dataset_stats}).

\vspace{-0.8em}
\paragraph{Real-World Test: Agentic Programming}
The agentic programming dimension is set as the first OOD validation of the C-A-F formulation, verifying whether routers developed on the probing set can generalize to fundamentally new task types. These tasks require multi-step planning, file navigation, and iterative debugging, which are qualitatively different from the 9 single-turn coding dimensions. The initial version involves 176 tasks extracted from SWE-bench Verified~\citep{openai_swebench_verified}, LongCLI-Bench~\citep{feng2026longcli}, FeatureBench~\cite{zhou2026featurebench}, and SWE-CI~\cite{chen2026swe}, by filtering tasks with high similarity with probing set. Evaluation uses Docker-based sandbox execution. Specifically, we use mini-swe-agent\footnote{https://github.com/SWE-agent/mini-swe-agent} 
with the SWE-Bench Docker harness~\cite{jimenez2024swebench}.
Construction details are provided in the Appendix~\ref{app:split}.


\subsection{Model Pool}

\begin{figure*}[t]
\centering
\includegraphics[width=\textwidth]{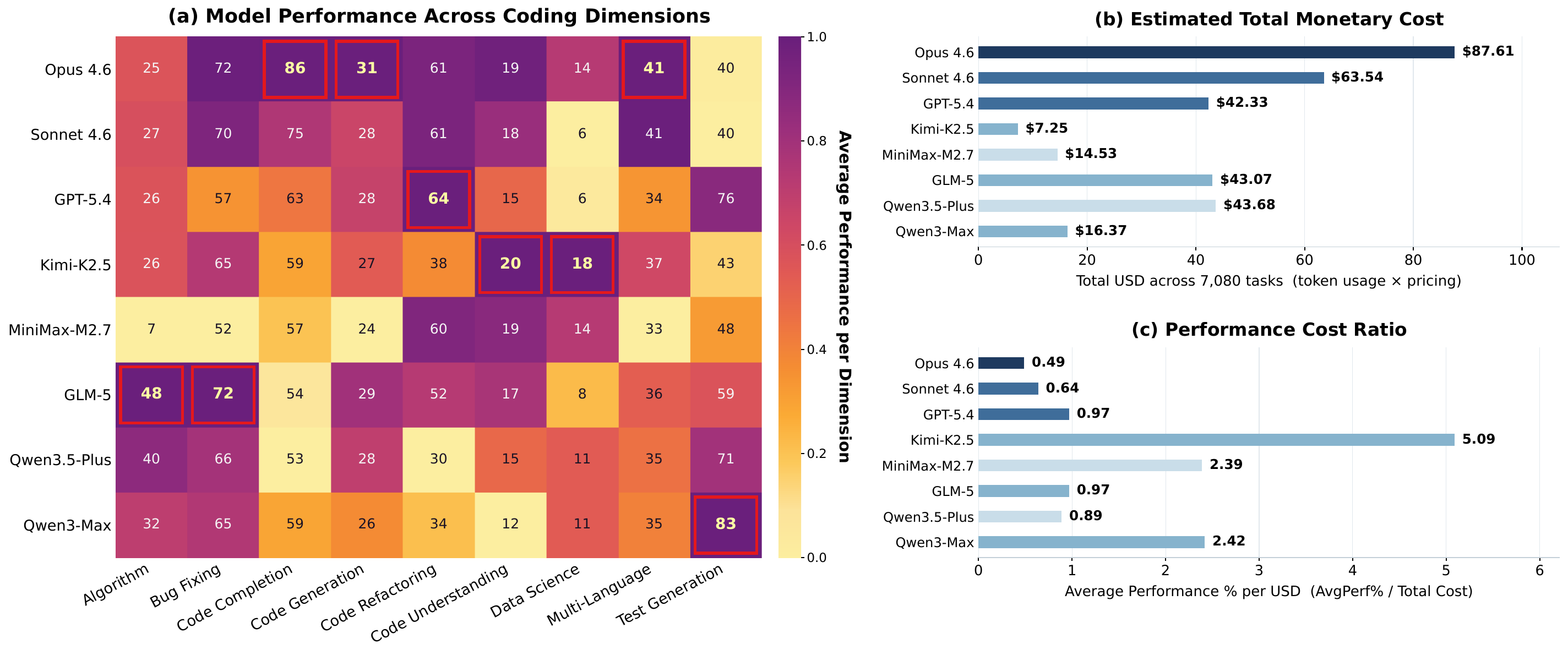}
\vspace{-2.2em}
\caption{\textbf{Performance, cost, and efficiency analysis.} (a) Performance heatmap of 8 models across 9 coding dimensions in the probing set, demonstrating that the optimal model varies significantly by task domain. (b) Total USD cost estimation over 7,080 probing set tasks. (c) Performance-to-cost ratio (AvgPerf\% / Total Cost). Claude Opus 4.6 has roughly $12\times$ the total cost of Kimi-K2.5.}
\vspace{-1.7em}
\label{fig:complementarity}
\end{figure*}

We evaluate eight frontier LLMs: Claude Opus 4.6 and Claude Sonnet 4.6~\cite{anthropic_opus46,anthropic_sonnet46}, GPT-5.4~\cite{openai_gpt54}, Qwen3-Max~\cite{yang2025qwen3} and Qwen3.5-Plus~\cite{qwen35_blog}, GLM-5~\cite{zai_glm5}, Kimi-K2.5~\cite{moonshot_kimi25}, and MiniMax-M2.7~\cite{minimax_m27}. The observation matrix of per-dimension performance is shown in Fig.~\ref{fig:complementarity}. 

We find that \textbf{no single model dominates across all coding dimensions.} Claude Opus 4.6 achieves the highest average (42.9\%) but is outperformed on algorithm design by GLM-5 (47.2\% vs.\ 25.4\%, an 86\% relative improvement), on test generation by Qwen3-Max (82.7\% vs.\ 39.2\%, a 111\% relative improvement), and on data science by Kimi-K2.5 (18.4\% vs.\ 14.2\%, a 30\% improvement). 5 distinct models serve as the dimension-best choice across the 9 dimensions and the costs of stronger models tend to be higher, confirming the value of model routing for both performance and cost efficiency.

\section{Empirical Validation}
\label{sec:experiments}

\subsection{Metrics}
\label{sec:metrics}

\textbf{AvgPerf}: average performance score across all tasks.
\textbf{CumReg}: the terminal cumulative regret calculated by Eq.~\ref{eq:regret} ($\epsilon_1,\epsilon_2=1,-0.1$). 
TotTok: total input and output token consumption (router$+$model). \$Total: calculated USD based on TotTok and official pricing (Appendix~\ref{app:pricing}). For local GPU-served routers (Finetuned LLMs, etc.), tokens are priced at \$0.054/M (see Appendix~\ref{app:local_price} for derivation). \textbf{Perf/\$}: performance-to-cost ratio ($\text{AvgPerf\%} / \text{Cost}$). Above metrics are computed across all tasks.


\definecolor{oursbg}{HTML}{ADE7F6}
\definecolor{boundbg}{HTML}{E3F2FD}

\begin{table}[!t]
\caption{Routing results, grouped by component-configuration taxonomy. \textbf{Left}: in-distribution test across 9 single-turn coding dimensions. \textbf{Right}: real-world OOD test on agentic programming. \textbf{CumReg}: cumulative regret across all tasks. \textbf{Perf/\$}: AvgPerf\% per USD. 
DimensionBest is not applicable to OOD test as it requires predefined dimension-to-model mapping, which is unavailable for unseen agentic-programming tasks.
The full breakdown is in the appendix.
}
\vspace{-0.2em}
\label{tab:main}
\centering
\small
\setlength{\tabcolsep}{4.5pt}
\renewcommand{\arraystretch}{1.05}
\begin{tabular}{@{}l l ccc ccc@{}}
\toprule
& & \multicolumn{3}{c}{\textbf{In-Distribution ($n{=}2{,}919$)}} & \multicolumn{3}{c}{\textbf{OOD Test (n=176)}} \\
\cmidrule(lr){3-5} \cmidrule(lr){6-8}
& \textbf{Router}
& \textbf{AvgPerf\%}$\uparrow$ & \textbf{CumReg}$\downarrow$ & \textbf{Perf/\$}$\uparrow$
& \textbf{AvgPerf\%}$\uparrow$ & \textbf{CumReg}$\downarrow$ & \textbf{Perf/\$}$\uparrow$ \\
\midrule
\rowcolor{boundbg}
& Oracle & 57.00 & 0 & 8.20 & 75.89 & 0 & 2.32 \\
\midrule
\multicolumn{8}{l}{\textbf{Agent-as-a-Router}} \\
\cmidrule(l){1-8}
& ACRouter (ours) & \textbf{49.98} & \textbf{205.5} & 3.79 & \textbf{62.50} & \textbf{17.0} & 1.18 \\
\midrule
\multicolumn{8}{l}{\textbf{Dynamic: Online Bandit}} \\
\cmidrule(l){1-8}
& LinTS & 46.48 & 307.4 & 4.49 & 46.43 & 35.9 & 0.75 \\
& LinUCB & 46.84 & 296.9 & 4.38 & 49.82 & 31.1 & 0.96 \\
\midrule
\multicolumn{8}{l}{\textbf{Static: Heuristic}} \\
\cmidrule(l){1-8}
& DimensionBest & 47.50 & 277.4 & 3.69 & --- & --- & --- \\
& kNN Retrieval & 47.18 & 286.7 & 6.07 & 14.29 & 66.7 & \textbf{1.45} \\
\midrule
\multicolumn{8}{l}{\textbf{Static: Trained Policy}} \\
\cmidrule(l){1-8}
& LogReg & 47.26 & 284.4 & 6.27 & 19.64 & 61.8 & 1.17 \\
& RouteLLM-BERT & 47.22 & 285.5 & 6.22 & 21.43 & 59.4 & 1.30 \\
& TF-IDF+MLP & 46.97 & 292.8 & 6.11 & 13.39 & 67.9 & 1.17 \\
& Qwen3.5-0.8B-Finetuned & 46.41 & 309.1 & 6.82 & 55.36 & 27.2 & 0.74 \\
& RouteLLM-MF & 46.16 & 316.5 & 6.19 & 8.93 & 72.7 & 0.94 \\
\midrule
\multicolumn{8}{l}{\textbf{Single-Model Baselines}} \\
\cmidrule(l){1-8}
& Always-Opus 4.6 & 43.83 & 387.1 & 1.29 & 57.14 & 26.7 & 0.64 \\
& Always-Kimi-K2.5 & 36.66 & 593.3 & \textbf{12.62} & 18.75 & 62.3 & 1.22 \\
& Always-Qwen3.5-Plus & 37.16 & 580.2 & 2.05 & 2.68 & 80.1 & 0.19 \\
& Random & 38.75 & 533.6 & 2.48 & 31.25 & 50.4 & 0.85 \\
\bottomrule
\end{tabular}
\vspace{-2em}
\end{table}


\subsection{Main Results and Observations}
\label{sec:main_results}

\textbf{ACRouter achieves the best AvgPerf among the routers in Table~\ref{tab:main} on both ID and OOD tasks by dynamically accumulating context, with lower cost than always choosing the strongest single model Opus-4.6}. Table~\ref{tab:main} reports the routing comparison on both the ID test tasks (left, $n{=}2{,}919$) and the OOD agentic programming tasks (right). On the ID test, ACRouter reaches the highest AvgPerf (49.98\%) and the lowest cumulative regret (205.5), beating DimensionBest with a full dimension-level prior by 2.48\% AvgPerf. On the OOD test, it reaches 62.50\% AvgPerf, ahead of other routers in Table~\ref{tab:main}, including the strongest single model strategy Always-Opus (57.14\%) and the finetuned Qwen3.5-0.8B (55.36\%). An updated standalone GPT-5.4 backend run resolves 75.00\% of the same OOD split (Appendix~\ref{app:ood_breakdown}), showing that this OOD setting also exposes strong backend-level gains beyond the original single-model baselines. ACRouter still achieves higher cost-efficiency than always choosing Opus: ACRouter's Perf/\$ is 3.79 on ID and 1.18 on OOD, both exceeding Always-Opus (1.29 ID, 0.64 OOD). 


\textbf{Static learners reach fair AvgPerf on the in-distribution test within the same distribution but generalize poorly on OOD tasks (lower AvgPerf than Always-Opus 4.6).} Table~\ref{tab:main} also evaluates each router on OOD tasks. These OOD agentic programming tasks are more representative of real-world settings, sharing minimal overlap with the 9 single-turn coding dimensions used to calibrate the routers. The lightweight classifiers (LogReg, TF-IDF+MLP, RouteLLM-MF, RouteLLM-BERT) remain within 1.3\% AvgPerf gap compared with DimensionBest on the in-distribution test, but their OOD AvgPerf drops sharply to 8.93\%--21.43\%, even lower than Random (31.25\%). This suggests that these routers noticeably overfit to the distribution of the training set, making them hard to generalize under a substantial distribution shift. 

\begin{figure}[t]
\centering
\includegraphics[width=0.98\textwidth]{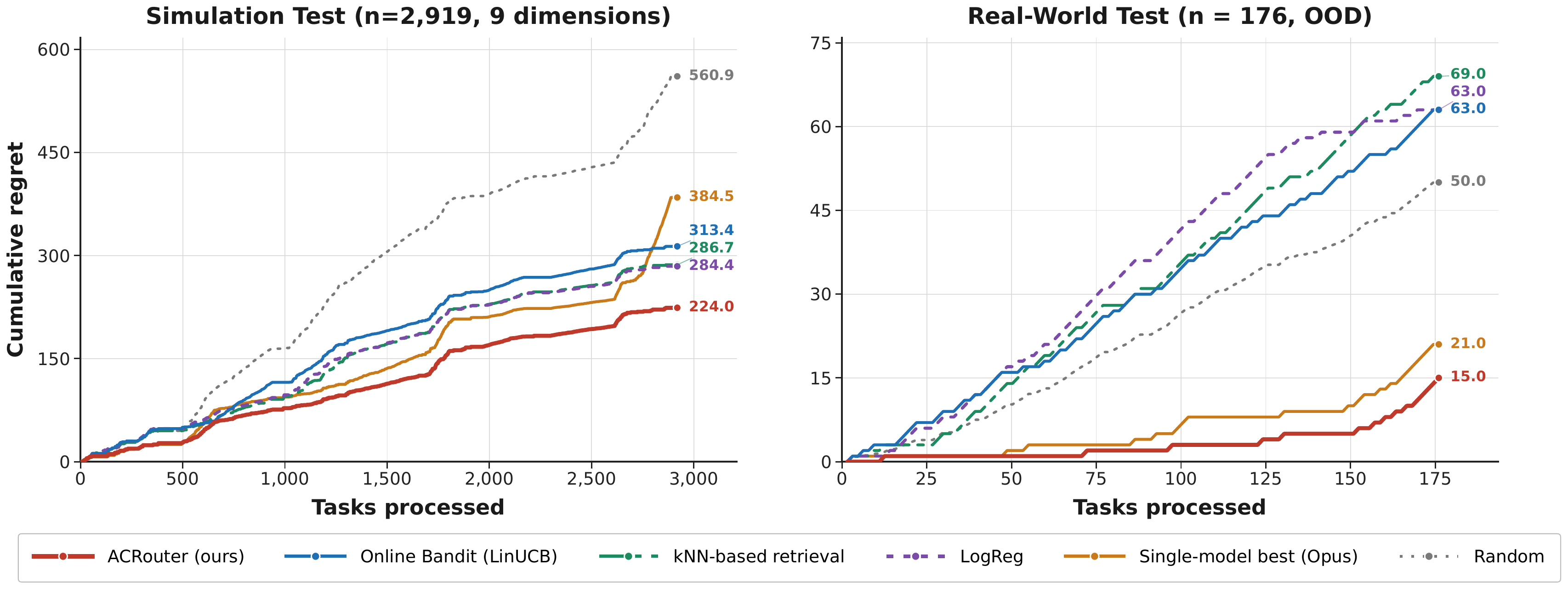}
\vspace{-0.5em}
\caption{\textbf{Cumulative regret across task streams.} Static routers grow faster on the in-distribution test (Left) and collapse on OOD real-world tasks (Right), while ACRouter shows lower regret as its Memory module accumulates verified experience.}
\label{fig:regret}
\end{figure}

\begin{figure}[t]
\centering
\includegraphics[width=0.99\textwidth]{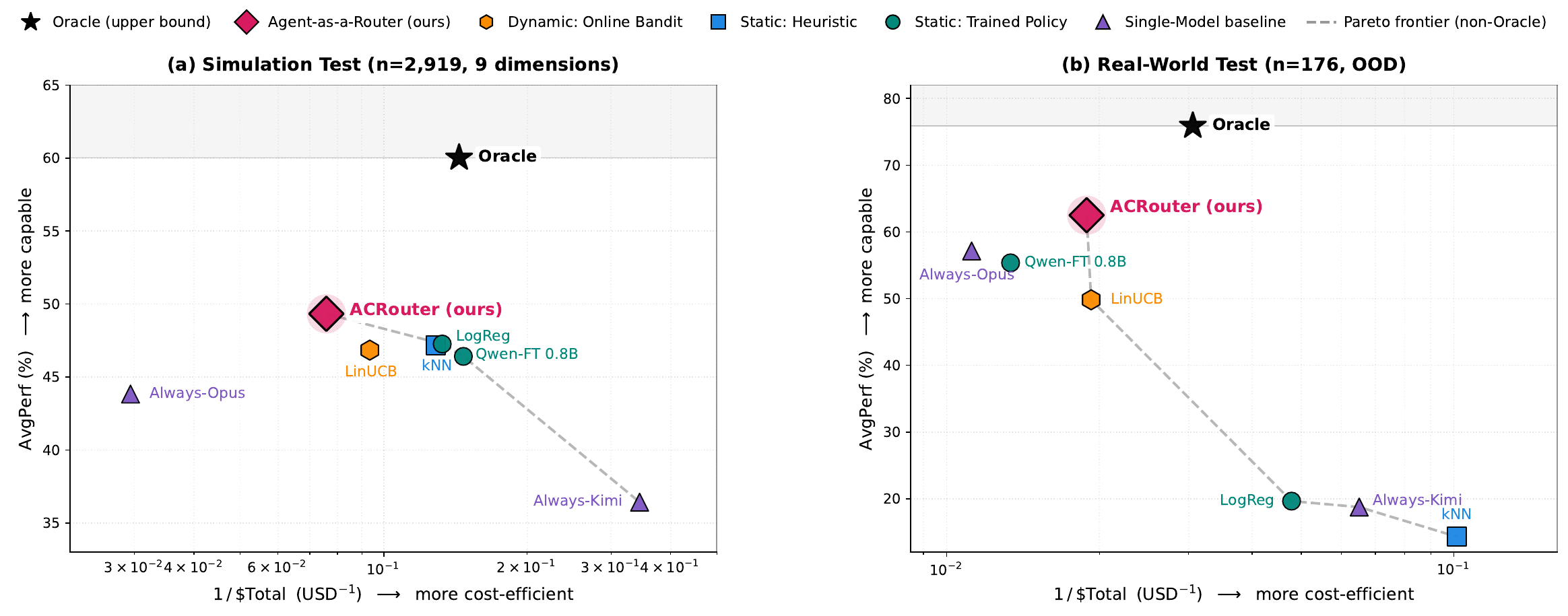}
\vspace{-0.8em}
\caption{\textbf{Cost--performance Pareto frontier analysis.} The dashed line traces the optimal trade-off for deployable routers. ACRouter extends the frontier upward in both ID and OOD with the highest AvgPerf and less cost than always choosing a premium model like Always-Opus.}
\label{fig:pareto}
\vspace{-1em}
\end{figure}

We note that contextual bandits (LinUCB, LinTS) keep updating online and also survive better, reaching 49.82\% / 46.43\% OOD, but they still lag behind ACRouter on both AvgPerf and CumReg because their per-arm linear models lack the context-aware reasoning that Orchestrator and Memory provide. Fig.~\ref{fig:regret} confirms the ranking: static methods tend to accumulate higher regret (284--317 for lightweight classifiers), bandits trail slightly (297--307), and only ACRouter (205.5) shows lower regret as Memory fills the information gap during deployment. Fig.~\ref{fig:pareto} also traces the Pareto frontier across all routers. ACRouter pays a higher cost (Perf/\$ 3.79) for Memory and Verifier but sits above the router frontier on AvgPerf (49.98 ID, 62.50 OOD), while the updated standalone GPT-5.4 OOD result is reported separately in Appendix~\ref{app:ood_breakdown}.







\subsection{Discussion}

\paragraph{Takeaways.} We summarize the four findings of the paper:
(1) the main performance bottleneck of an LLM router is \emph{information deficit} rather than reasoning failure;
(2) no single model dominates all coding dimensions;
(3) full Agent-as-a-Router achieves the best performance on both in-distribution and OOD tasks, and still costs less than always choosing Opus-4.6;
(4) static learners achieve fair performance on in-distribution tasks but fail significantly on OOD tasks, e.g., lightweight classifiers like RouteLLM-BERT achieve fair Perf/\$ on in-distribution tasks but barely handle OOD tasks.

\paragraph{Limitations.}
A key limitation is that provider-side cache hit rates are not observable, so our monetary cost estimates are based on public token prices and measured token usage.
\emph{Therefore, monetary cost serves only as a secondary reference metric for relative comparison rather than acting as a primary indicator.}
The agentic programming evaluation uses a 40-step limit instead of the standard 250-step configuration to keep the budget tractable (the relative router comparison is unaffected). Our C-A-F instantiation uses an LLM policy with a memory-kNN ensemble; alternative instantiations (e.g., advanced parameter-level memory techniques) remain to be explored.

\section{Conclusion}
\label{sec:conclusion}
In this work, we propose \textbf{Agent-as-a-Router}, a routing framework that acquires execution-grounded information through a Context-Action-Feedback loop, naturally formalizable as a contextual bandit with cumulative regret as its streaming metric.
We instantiate the framework as \textbf{ACRouter}, composed of an Orchestrator, a Verifier, and a Memory module, and evaluate it on \textbf{CodeRouterBench}, an evaluation environment built specifically to enable regret-based router comparison on streaming tasks. ACRouter attains the lowest cumulative regret on the in-distribution stream tasks and is the only router that maintains strong performance under
the OOD agentic-programming setting. More broadly, actively closing the information gap through execution-grounded feedback emerges as a general principle for building agentic systems that must route among heterogeneous tools or models.

{
    \bibliographystyle{bibstyle}
    \bibliography{references}
}

\newpage
\appendix

\section{ACRouter System Architecture}
\label{sec:agent_arch}

\textbf{ACRouter} (\S~\ref{sec:acrouter}) is composed of a two-layer modular architecture: a \emph{decision layer} of three core modules that realize the C-A-F loop, and a \emph{tool layer} of shared infrastructure that the core modules invoke (see Fig.~\ref{fig:agent_design}). This appendix details the architecture (\S\ref{app:arch_modules}), the reward-weight conventions used throughout the paper (\S\ref{app:arch_reward}), and the C-A-F configuration taxonomy that partitions all reported routing methods (\S\ref{app:arch_taxonomy}).

\subsection{Core Modules and Tool Layer}
\label{app:arch_modules}

\paragraph{Core modules (decision layer).}
\begin{itemize}[nosep]
\item \textbf{Orchestrator}: the central coordinator that selects which candidate model to invoke for the current task $t_i$. It consults Memory state, ingests the DimensionBest prior, the top-$10$ historical neighbors retrieved from Memory by cosine kNN, and the task metadata, and uses a fine-tuned Qwen3.5-0.8B policy combined with heuristic rules via weighted voting to make the final dispatch decision.
\item \textbf{Verifier}: a sandbox-native confidence estimator that examines each candidate's output and aggregates multiple signal tiers (AST parse, sandbox execution, prompt-embedded tests, and rule-based signals) into a unified score $u_i\in[0,1]$ via Eq.~\ref{eq:verifier}. The Verifier produces the verdict that is written into Memory.
\item \textbf{Memory}: an online vector store keyed by task embeddings (voyage-code-3 / BGE-large) whose value logs the chosen model, observed performance, monetary cost, and the Verifier's verification trace. Retrieval uses cosine kNN over embeddings (similarity threshold $0.5$, $k{=}10$). The store is FIFO-bounded at 20K entries and is committed in place after each loop. Per-task granularity makes both past successes and recent failures of any candidate on similar tasks visible to the Orchestrator at the next decision.
\end{itemize}

\paragraph{Tool layer (execution infrastructure).} Shared resources invoked by the core modules; they are not independently comparable strategies but are essential for instantiating any router under the C-A-F loop.
\begin{itemize}[nosep]
\item \textbf{Candidate Models} (the model pool): the 8 LLMs available for routing (Claude Opus~4.6, Sonnet~4.6, GPT-5.4, Qwen3-Max, Qwen3.5-Plus, Kimi-K2.5, GLM-5, MiniMax-M2.7). The Orchestrator selects from these.
\item \textbf{Routing Tools} (serve the Orchestrator): dimension-best lookup tables, trained classifiers, and LLM-based selectors. These are the concrete mechanisms the Orchestrator uses to make its selection.
\item \textbf{Evaluation Tools} (serve the Verifier): AST parser, sandbox runner, self-consistency checker (k-sample output agreement), LLM-as-Judge, and a prompt-test extractor that identifies in-prompt test cases. All produce quality signals \emph{without} ground-truth oracle tests; the Verifier aggregates them into a confidence score.
\item \textbf{Embedding Encoder} (serves Memory): maps a task's prompt text into a dense vector for kNN retrieval. Implementations can range from a code-specialized API (e.g., voyage-code-3) to a local open-source model (e.g., BGE-large).
\item \textbf{Infrastructure}: execution sandbox (Docker, timeouts), context parser (extracts dimension/difficulty/language from raw input), and an online updater for Memory statistics.
\end{itemize}

\subsection{Per-Loop Reward Weights and Reported Regret}
\label{app:arch_reward}

Eq.~\ref{eq:reward1} defines the per-task cost-aware reward $r_i(a_i) = \epsilon_1\,s_i(a_i) + \epsilon_2\,\kappa_i(a_i)$ with $\epsilon_1 > 0$ and $\epsilon_2 < 0$. Throughout the paper, AvgPerf reports the raw mean score, while CumReg uses the canonical evaluation reward $(\epsilon_1,\epsilon_2)=(1,-0.1)$:
\[
  r_i(a_i)=s_i(a_i)-0.1\,\kappa_i(a_i),
\]
where $\kappa_i$ is measured in USD under Table~\ref{tab:pricing}. On the in-distribution single-turn split this cost term is mostly a soft tie-breaker; on OOD agentic tasks, multi-step backend calls make the same cost term materially affect regret. \$Total and Perf/\$ are still reported as separate columns to expose absolute deployment cost.

\paragraph{Per-task (micro) oracle.}
The reference oracle in Eq.~\ref{eq:oracle} computes $r^*_i = \max_{j \in [M]}\, R_{ij}$ \emph{independently for each task}, where $R_{ij} = \epsilon_1\,s_{ij} + \epsilon_2\,\kappa_{ij}$ is the cost-aware reward of model $j$ on task $i$. The cumulative regret in Table~\ref{tab:main} therefore measures the per-task gap
\begin{equation*}
  \mathrm{CumReg}_N(\pi) \;=\; \sum_{i=1}^{N} \bigl(r^*_i - r_i(a_i)\bigr) \;=\; \sum_{i=1}^{N} \Bigl[\,\max_{j \in [M]} R_{ij} \;-\; R_{i,\,a_i}\Bigr].
\end{equation*}
This metric is not the gap to any single-best-arm policy that commits to one global model, nor the gap to the oracle's average score alone. The oracle is recomputed per task from the complete observation matrix, and CumReg sums those task-level reward gaps.

\paragraph{Bandit reward.}
The online bandit baselines (LinUCB, LinTS) use $(\epsilon_1, \epsilon_2) = (1, -0.1)$ for their per-arm posterior updates as stated in \S\ref{sec:variants}, matching the standard contextual-bandit cost-aware formulation~\citep{li2010contextual,agrawal2013thompson}. This is independent of the $(1, -0.1)$ weights used to \emph{evaluate} CumReg in Table~\ref{tab:main}: every router (bandit, classifier, ACRouter) is scored under the same canonical evaluation reward, so the column is directly comparable across method families.

\subsection{Configuration Taxonomy}
\label{app:arch_taxonomy}

Different routing strategies correspond to different subsets of the C-A-F loop being active (Table~\ref{tab:components}). This decomposition mirrors \S\ref{sec:variants} and structures the ablation in \S\ref{sec:experiments}.

\begin{table}[h]
\caption{Routing methods organized as C-A-F configurations. Loop-broken methods leave Memory empty or static (frozen probing prior); the loop-complete ACRouter activates all three core modules.}
\label{tab:components}
\centering
\small
\resizebox{\textwidth}{!}{%
\begin{tabular}{lccc}
\toprule
\textbf{Method family} & \textbf{Orchestrator} & \textbf{Verifier} & \textbf{Memory} \\
\midrule
Single-Model (\emph{Always-$m$})              & Direct dispatch       & ---                  & ---                          \\
Static: Heuristic (DimensionBest, kNN Retrieval) & Static lookup     & ---                  & Frozen probing-set prior     \\
Static: Trained Policy (LogReg / TF-IDF+MLP / RouteLLM / Qwen3.5-FT) & Trained model  & ---  & ---                  \\
Dynamic: Online Bandit (LinUCB, LinTS)        & $\arg\max$ rule       & Reward only          & Per-arm parametric posterior \\
\textbf{ACRouter (loop-complete)}             & LLM policy + tools    & Sandbox-native       & Online task-embedding kNN    \\
\bottomrule
\end{tabular}%
}
\end{table}

\section{Benchmark and Setup Details}
\label{app:benchmark}

This section consolidates the construction details of CodeRouterBench: the coding dimensions and their sources (\S\ref{app:dims}), the deterministic 70/30 split protocol (\S\ref{app:split}), the API pricing and self-hosted serving rate used for every dollar number reported (\S\ref{app:pricing}), and the prompt templates used by the LLM-as-a-Router baselines and the +Perf~stats ablation (\S\ref{app:prompts}).

\subsection{Coding Dimensions}
\label{app:dims}

\begin{table}[h]
\caption{Coding dimensions in CodeRouterBench. The 9 single-turn dimensions each contain 1{,}111 tasks ($70/30$ deterministic MD5 split into probing and in-distribution test, see Section~\ref{app:split}). The 10th dimension, agentic programming, is held out as the OOD test (176 tasks). Eval: Exec~=~execution-based pass@1; Proxy+~=~proxy metric supplemented by LLM-as-Judge; Sandbox~=~Docker-based patch application and test execution.}
\label{tab:dimensions}
\centering
\small
\resizebox{\textwidth}{!}{%
\begin{tabular}{llll}
\toprule
\textbf{Dimension} & \textbf{Description} & \textbf{Source(s)} & \textbf{Eval} \\
\midrule
Code Generation     & Function-level synthesis        & HumanEval+, MBPP+, BigCodeBench       & Exec   \\
Algorithm Design    & Competitive programming         & LiveCodeBench, BigCodeBench           & Exec   \\
Bug Fixing          & Locate and repair defects       & DebugBench, SWE-bench Lite            & Exec   \\
Code Completion     & Fill-in-the-middle              & CRUXEval, HumanEval+ variants         & Exec   \\
Code Refactoring    & Improve code quality            & Bugs2Fix, CanItEdit                   & Proxy+ \\
Data Science        & Data analysis pipelines         & DS-1000, BigCodeBench                 & Exec   \\
Multi-Language      & Cross-language tasks            & HumanEval-X, MultiPL-E                & Exec   \\
Code Understanding  & Explain \& summarize code       & CodeXGLUE Summarization               & Proxy+ \\
Test Generation     & Generate test suites            & LiveCodeBench, HumanEval+ variants    & Proxy+ \\
\midrule
Agentic Programming (OOD) & Long-horizon, multi-file repository tasks & SWE-bench Verified, LongCLI-Bench, FeatureBench, SWE-CI & Sandbox \\
\bottomrule
\end{tabular}%
}
\end{table}

\subsection{Data Split and Evaluation Levels}
\label{app:split}

The 9 single-turn dimensions are split deterministically (md5-based, seed \texttt{coding-router-v1}) into three roles that mirror a typical router development workflow:

\begin{itemize}[nosep]
\item \textbf{Probing set} (train 60\% + val 10\% = 7{,}080 tasks across 9 single-turn dimensions): Used to \emph{develop} routers --- profile model strengths per dimension, train classifiers, calibrate DimensionBest, and warm-start ACRouter's Memory module (200 val tasks). This is the dev split for building custom routers.
\item \textbf{In-distribution test} (test 30\% = 2{,}919 tasks across 9 single-turn dimensions): Held-out evaluation in a controlled environment with execution-verified per-task per-model scores. All in-distribution (ID) numbers in Table~\ref{tab:main} (left) use this split. The ID per-dim task counts (306--347 per dimension) sum to 2{,}919.
\item \textbf{OOD test} (Real-world agentic programming, 176 tasks): Held out as the agentic-programming dimension. No router has access to any ground-truth data from this dimension. Tests whether a router developed on the probing set generalizes to fundamentally different task types. Tasks are extracted from SWE-bench Verified~\citep{openai_swebench_verified}, LongCLI-Bench~\citep{feng2026longcli}, FeatureBench~\citep{zhou2026featurebench}, and SWE-CI~\citep{chen2026swe}, filtering out tasks with high prompt similarity to the probing set. Evaluation uses the SWE-Bench Docker harness~\citep{jimenez2024swebench} with mini-swe-agent, capped at 40 steps per task.
\end{itemize}

\subsection{Model Pricing}
\label{app:pricing}

Backend tokens are priced at official API rates (\S\ref{app:api_pricing}); self-hosted router-side tokens use a measured H100-amortised rate of \$0.054/M (\S\ref{app:local_price}).

\subsubsection{API Pricing}
\label{app:api_pricing}

Table~\ref{tab:pricing} presents the per-model API pricing used for cost calculations throughout the paper. All values are mirrored verbatim from \texttt{configs/model\_pricing.json} in the released artifact, which is the single source of truth for backend pricing in every method's recompute.

\begin{table}[h]
\caption{Per-model API pricing in USD per million tokens (input/output asymmetric), used in every cost number reported throughout the paper. The 8 models are partitioned into tiers (premium / high / mid / low) ordered by their geometric mean of input and output prices.}
\label{tab:pricing}
\centering
\small
\begin{tabular}{lccc}
\toprule
\textbf{Model}      & \textbf{\$/M input} & \textbf{\$/M output} & \textbf{Tier} \\
\midrule
Claude Opus 4.6     & \$5.00              & \$25.00              & premium \\
Claude Sonnet 4.6   & \$3.00              & \$15.00              & high    \\
GPT-5.4             & \$2.50              & \$15.00              & high    \\
Qwen3-Max           & \$1.20              & \$6.00               & mid     \\
GLM-5               & \$0.88              & \$3.22               & mid     \\
Kimi-K2.5           & \$0.60              & \$3.07               & mid     \\
Qwen3.5-Plus        & \$0.40              & \$2.40               & low     \\
MiniMax-M2.7        & \$0.30              & \$1.20               & low     \\
\bottomrule
\end{tabular}
\end{table}

\subsubsection{Local Serving Token Cost Calculation}
\label{app:local_price}

Self-hosted routers (the orchestrator LLM in ACRouter and the LoRA-finetuned Qwen3.5 routers) do not incur a per-token API charge. We instead amortise GPU rental cost over the measured serving throughput, yielding a single combined-throughput price in USD per million tokens. All numbers below are reproducible from the artifact released alongside the paper (\texttt{local\_pricing\_benchmark/}).

\paragraph{Hardware and pricing.}
Throughput is measured on a single NVIDIA H100 80GB HBM3, the deployment target used by every self-hosted router experiment in this paper. We adopt a representative on-demand H100 rental rate of \$6.88/GPU-hour as the cost basis (CoreWeave\footnote{https://www.coreweave.com/pricing} / Lambda Cloud\footnote{https://aws.amazon.com/lambda/pricing/} public pricing at the time of writing).

\paragraph{Workload.}
We replay the 2{,}919-task in-distribution test split as router queries: each input is the actual zero-shot router prompt (system message + 8-model description + the task's dimension and prompt, truncated to 2{,}000 characters), and the model is asked to emit a JSON model-pick. This matches the production routing call that finetuned Qwen variants make in our experiments, so the measured throughput is representative of real router-side load rather than a synthetic best case.

\paragraph{Configuration.}
We use the offline engine of SGLang~\cite{zheng2024sglang} (no HTTP overhead) running Qwen3.5-0.8B in bfloat16 with flashInfer attention. Decoding is greedy ($T{=}0$) with \texttt{max\_new\_tokens}$=96$ and a stop sequence on three consecutive newlines, matching the JSON-structured output expected by the router parser. Requests are issued in batches of $64$ and looped over the 2{,}919 prompts in a fixed seed-42 shuffle. We discard the first $32$ warm-up prompts (kernel compilation, KV-cache pre-allocation) and then time exactly 300.5 seconds of steady-state generation.

\paragraph{Measured throughput.}
Over the 300.5-second window the engine processed $22{,}848$ requests, consuming $8{,}393{,}247$ input tokens and producing $2{,}152{,}601$ output tokens, for a combined sustained throughput of:
\begin{equation}
\mathrm{TPS}_{\text{in}+\text{out}} \;=\; \frac{8{,}393{,}247 + 2{,}152{,}601}{300.5\,\text{s}} \;\approx\; 35{,}094 \text{ tokens/s}.
\end{equation}
Splitting the throughput by direction yields $27{,}931$ input-tokens/s and $7{,}163$ output-tokens/s; the routing prompt is input-heavy (4:1 input:output) because the model emits only a short JSON pick.

\paragraph{Derived per-token cost.}
Multiplying by 3{,}600 yields $\sim 1.263 \times 10^{8}$ tokens per GPU-hour, so the rental cost per million combined tokens is:
\begin{equation}
\frac{\$6.88}{1\text{ hour}} \cdot \frac{1\text{ hour}}{1.263 \times 10^{8}\text{ tokens}} \cdot 10^{6}
\;=\; \boxed{\$0.054 \text{ per }1\text{M tokens}}.
\label{eq:local_price}
\end{equation}
This is the rate used for self-hosted router-side tokens of ACRouter and the Finetuned Qwen3.5-0.8B router throughout the paper. Backend model tokens are always priced at API rates (Table~\ref{tab:pricing}); only the orchestrator/router LLM that we run ourselves is amortized this way.

\paragraph{Reproducibility.}
The exact benchmark script, prompt-construction code, and resulting JSON (including per-15-second progress logs) will be released. Re-running on a different H100 with SGLang and the published Qwen3.5-0.8B snapshot reproduces the throughput within $\pm 5\%$.

\subsection{Router Prompt Templates}
\label{app:prompts}

Three prompt variants are explored for the supplementary experiments in LLM routers: a zero-shot Vanilla template (\S\ref{app:prompt_zero}, used in the main experiments for trained policy), a 3-shot in-context variant (\S\ref{app:prompt_few}), and a +Perf~stats ablation that injects probing-set per-dimension scores (\S\ref{app:prompt_perfstats}).

\subsubsection{Zero-Shot Router Prompt}
\label{app:prompt_zero}

The zero-shot LLM router (Vanilla in Table~\ref{tab:bias}) receives the following system prompt:

\begin{figure}[h]
\begin{tcolorbox}[
  colback=gray!5,
  colframe=gray!50,
  boxrule=0.4pt,
  arc=2pt,
  left=6pt, right=6pt, top=4pt, bottom=4pt,
  fontupper=\small\ttfamily,
  fonttitle=\small\sffamily\bfseries,
  coltitle=black,
  colbacktitle=gray!15
]
\small
\texttt{You are a coding task router. Your objective is to maximize the performance-cost trade-off: choose the model that achieves the best quality for its cost on this task. [8 models listed with capability descriptions, sorted by cost tier]. Prefer cheaper models when quality is comparable. Respond with JSON: \{"model": "...", "reasoning": "..."\}.}
\end{tcolorbox}
\caption{Prompt template for zero-shot router.}
\label{fig:coding-router}
\end{figure}

\subsubsection{Few-Shot Router Prompt}
\label{app:prompt_few}

The 3-shot LLM router uses the same system prompt as the zero-shot variant, but prepends three demonstration examples before the target task. The prompt structure is:

\begin{figure}[h]
\begin{tcolorbox}[
  colback=gray!5,
  colframe=gray!50,
  boxrule=0.4pt,
  arc=2pt,
  left=6pt, right=6pt, top=4pt, bottom=4pt,
  fontupper=\small\ttfamily,
  fonttitle=\small\sffamily\bfseries,
  coltitle=black,
  colbacktitle=gray!15
]
\small
\texttt{\#\# Examples}\\[2pt]
\texttt{\#\#\# Example 1}\\
\texttt{- Dimension: bug\_fixing}\\
\texttt{- Difficulty: medium}\\
\texttt{- Language: python}\\
\texttt{- Prompt: <first 300 chars of task prompt>...}\\
\texttt{- \textbf{Best model}: glm-5}\\
\texttt{- Scores: claude-opus-4-6: 0.72, claude-sonnet-4-6: 0.70, glm-5: 0.73, gpt-5.4: 0.68, kimi-k2.5: 0.65, ...}\\[4pt]
\texttt{\#\#\# Example 2}\\
\texttt{[...similar format...]}\\[4pt]
\texttt{\#\#\# Example 3}\\
\texttt{[...similar format...]}\\[4pt]
\texttt{---}\\
\texttt{Now route the following task:}\\
\texttt{\#\# Task to Route}\\
\texttt{**Dimension**: code\_completion}\\
\texttt{**Difficulty**: hard}\\
\texttt{**Language**: python}\\
\texttt{**Prompt**: <full task prompt>}
\end{tcolorbox}
\caption{Few-shot prompt template for model routing.}
\label{fig:coding-router}
\end{figure}

\noindent\textbf{Example selection strategy.}
Examples are drawn from the probing split with the following design choices:
\begin{enumerate}[leftmargin=*,nosep]
    \item \textbf{Same-dimension priority}: all 3 examples are sampled from tasks sharing the same dimension as the target task. If fewer than 3 same-dimension examples are available, the remaining slots are filled from other dimensions.
    \item \textbf{Non-trivial examples only}: tasks where all 8 models achieve identical scores are excluded, since they carry no routing signal.
    \item \textbf{Oracle labels}: each example shows the oracle-best model (the model with the highest score for that task, with ties broken by cost ascending then alphabetical) along with all 8 models' scores, giving the LLM both the answer and the full score distribution.
    \item \textbf{Prompt truncation}: task prompts in examples are truncated to 300 characters to control input token cost; the target task's prompt is included in full.
    \item \textbf{Fixed seed}: examples are sampled with a fixed random seed (42) for reproducibility.
\end{enumerate}

\noindent\textbf{Response format.}
The LLM is instructed to respond with a JSON object: \texttt{\{"model": "<name>", "reasoning": "<explanation>"\}}. Responses are parsed with a multi-strategy fallback: (1) direct JSON parse, (2) regex extraction from markdown code blocks, (3) model-name string matching, (4) fallback to the default model (\texttt{claude-sonnet-4-6}).

\subsubsection{Performance-Statistics Ablation Prompt}
\label{app:prompt_perfstats}

In the +Perf~stats ablation (Table~\ref{tab:bias}), model descriptions are replaced with tabular performance data drawn from the probing set:

\begin{figure}[h]
\begin{tcolorbox}[
  colback=gray!5,
  colframe=gray!50,
  boxrule=0.4pt,
  arc=2pt,
  left=6pt, right=6pt, top=4pt, bottom=4pt,
  fontupper=\small\ttfamily,
  fonttitle=\small\sffamily\bfseries,
  coltitle=black,
  colbacktitle=gray!15
]
\small
\texttt{Model capabilities (average scores per dimension): Claude Opus 4.6: code\_gen=0.315, algo=0.254, bug\_fix=0.717, completion=0.860, refac=0.607, ds=0.142, multi=0.408, underst=0.193, test\_gen=0.392. [... similar for all 8 models]}
\end{tcolorbox}
\caption{Supplied tabular performance information in +Perf setting of ablation study.}
\label{fig:coding-router}
\end{figure}

\noindent The Anonymized variant additionally rewrites every model identifier to Model-A through Model-H to remove any prior the LLM may carry about specific provider strengths.

\section{Supplementary Experiments}
\label{app:experiments}

This section reports two experimental studies that complement the main results: an LLM-as-a-Router sweep over all 8 candidate models in 0-shot, 3-shot, and online-agent modes (\S\ref{app:llm_routers}); and a Qwen-router parameter-scaling sweep that isolates the contribution of router capacity (\S\ref{app:scaling}).

\subsection{LLM-as-a-Router Across All 8 Models}
\label{app:llm_routers}

\subsubsection{Setup}
\label{app:llm_routers_setup}

To verify that the information-deficit diagnosis in Table~\ref{tab:bias} is not specific to Claude Sonnet 4.6, we evaluate every one of the candidate models in the role of an LLM-as-a-Router on the in-distribution test split. Each model is given the same Vanilla zero-shot or 3-shot prompt template (Section~\ref{app:prompts}) and instructed to pick a backend model for each task. Costs use the pricing in Table~\ref{tab:pricing} for backend tokens; the LLM router's own tokens are priced at the same API rates as its backend role (since these are API-served, not self-hosted). For \textbf{Panel B} the 0-shot router is augmented with a cosine-kNN Memory of 200 warm-up tasks and online-updated as the test stream proceeds (seed 42).

\begin{table}[t!]
\caption{Router-model comparison on the 2{,}919-task in-distribution test split. Dollar columns use the updated pricing in Table~\ref{tab:pricing}; CumReg uses the per-task cost-aware oracle with $r=s-0.1\,\kappa_{\mathrm{USD}}$. \textbf{Panel A}: LLM-as-a-Router with all 8 models in 0-shot and 3-shot settings. \textbf{Panel B}: Agent online --- each agent starts with empty memory, warms up on 200 val tasks, then routes the test stream while accumulating cosine-kNN experience under verifier feedback (seed=42). Rows ordered by AvgPerf within each panel.}
\label{tab:llm_routers}
\centering
\scriptsize
\setlength{\tabcolsep}{4.5pt}
\resizebox{0.75\textwidth}{!}{%
\begin{tabular}{llcccc}
\toprule
\textbf{Mode} & \textbf{Router LLM} & \textbf{AvgPerf\%} & \textbf{CumReg}$\downarrow$ & \textbf{\$Total} & \textbf{Perf/\$}$\uparrow$ \\
\midrule
\multicolumn{6}{l}{\textit{Panel A: LLM-as-a-Router (0-shot)}} \\
\midrule
0-shot & Qwen3.5-Plus       & 46.87 & 296.3 & \$13.06 & 3.59 \\
0-shot & Qwen3-Max          & 46.60 & 304.1 & \$11.82 & 3.94 \\
0-shot & Kimi-K2.5          & 46.29 & 313.3 & \$14.11 & 3.28 \\
0-shot & GPT-5.4            & 45.91 & 324.0 & \$10.26 & 4.48 \\
0-shot & GLM-5              & 45.42 & 338.7 & \$14.56 & 3.12 \\
0-shot & MiniMax-M2.7       & 42.11 & 435.6 & \$16.90 & 2.49 \\
0-shot & Claude Sonnet 4.6  & 41.41 & 456.4 & \$21.02 & 1.97 \\
0-shot & Claude Opus 4.6    & 39.27 & 518.9 & \$21.20 & 1.85 \\
\midrule
\multicolumn{6}{l}{\textit{Panel A (cont.): LLM-as-a-Router (3-shot)}} \\
\midrule
3-shot & GLM-5              & 46.03 & 320.8 & \$12.22 & 3.77 \\
3-shot & Qwen3-Max          & 45.80 & 327.7 & \$14.63 & 3.13 \\
3-shot & GPT-5.4            & 45.66 & 331.3 & \$9.80  & 4.66 \\
3-shot & Claude Sonnet 4.6  & 45.53 & 335.3 & \$12.01 & 3.79 \\
3-shot & Qwen3.5-Plus       & 45.39 & 339.5 & \$12.64 & 3.59 \\
3-shot & MiniMax-M2.7       & 45.29 & 342.5 & \$13.88 & 3.26 \\
3-shot & Kimi-K2.5          & 45.10 & 348.5 & \$17.98 & 2.51 \\
3-shot & Claude Opus 4.6    & 41.17 & 463.5 & \$21.34 & 1.93 \\
\midrule
\multicolumn{6}{l}{\textit{Panel B: Agent online (200-task warm-up, seed=42)}} \\
\midrule
Agent & MiniMax-M2.7        & 45.27 & 343.2 & \$15.30 & 2.96 \\
Agent & GLM-5               & 45.21 & 345.0 & \$15.84 & 2.85 \\
Agent & Qwen3-Max           & 44.92 & 353.7 & \$18.00 & 2.50 \\
Agent & Qwen3.5-Plus        & 43.98 & 381.0 & \$16.17 & 2.72 \\
Agent & Claude Sonnet 4.6   & 43.65 & 390.7 & \$16.84 & 2.59 \\
Agent & Kimi-K2.5           & 42.70 & 417.9 & \$12.45 & 3.43 \\
Agent & GPT-5.4             & 40.95 & 468.9 & \$11.24 & 3.64 \\
\midrule
\emph{Reference}      & \emph{Random}        & \emph{38.75} & \emph{533.6} & \emph{\$15.64} & \emph{2.48} \\
\bottomrule
\end{tabular}%
}
\end{table}

\subsubsection{Results and Findings}
\label{app:llm_routers_findings}

\textbf{Three findings} stand out from Table~\ref{tab:llm_routers}. First, \emph{coding ability $\neq$ routing ability}: all 8 LLM routers remain below DimensionBest, and Claude Opus 4.6 ranks last in both 0-shot and 3-shot despite being the strongest individual coder. Second, few-shot prompting does not reliably solve routing: it slightly improves some router LLMs but still leaves substantial regret to the per-task oracle. Third, agent-mode online Memory is mixed across router LLMs; it helps MiniMax-M2.7 but does not uniformly improve stronger prompt-following routers. Random and DimensionBest are reported as references.

\subsection{Qwen Router Scaling Sweep}
\label{app:scaling}

\subsubsection{Setup}
\label{app:scaling_setup}

We sweep five Qwen3.5 sizes ($0.8$B / $2$B / $4$B / $9$B / $27$B) under unified LoRA finetuning (\textbf{FT v4}: attention$+$MLP, $r{=}16$, $\alpha{=}32$, dropout $0.05$, identical across sizes for a controlled scaling curve), evaluated on the canonical $n{=}2{,}919$ in-distribution test split. Qwen3.5-0.8B-Finetuned is the variant reported in the Trained-Policy block of Table~\ref{tab:main}; the larger sizes are included here to verify that scaling does not change the takeaway.

\subsubsection{Results and Findings}
\label{app:scaling_findings}

\begin{table}[h]
\centering
\small
\setlength{\tabcolsep}{5pt}
\caption{Qwen router scaling sweep on the canonical $n{=}2{,}919$ in-distribution test split (FT~v4 LoRA, attention$+$MLP, $r{=}16$, $\alpha{=}32$). \textbf{Toks}: average input$+$output tokens per task (router$+$backend). \textbf{CumReg}: per-task cost-aware regret; it is reported only where archived per-task decisions are available. The 0.8B row is the Qwen3.5-0.8B-Finetuned entry in Table~\ref{tab:main}.}
\label{tab:scaling}
\resizebox{0.7\textwidth}{!}{%
\begin{tabular}{@{}l c c r r@{}}
\toprule
\textbf{Size} & \textbf{AvgPerf\%}$\uparrow$ & \textbf{Gap\%}$\downarrow$ & \textbf{Toks/task} & \textbf{CumReg}$\downarrow$ \\
\midrule
0.8B & 46.41 & 18.6 & 500 & 309.1 \\
2B   & 46.69 & 18.1 & 501 & -- \\
4B   & 46.21 & 18.9 & 492 & -- \\
9B   & 46.56 & 18.3 & 496 & -- \\
27B  & \textbf{46.74} & \textbf{18.0} & 495 & -- \\
\bottomrule
\end{tabular}%
}
\end{table}

\paragraph{Findings.}
\begin{itemize}[nosep]
\item \textbf{Finetuning is the gate, not size.} Without finetuning, base Qwen3.5 9B/27B/35B emit malformed routing tokens that the parser falls back to a single default model (\texttt{claude-sonnet-4-6}); their AvgPerf collapses to the always-Sonnet floor ($0.4131$). Once finetuned, every size lifts AvgPerf to $\geq 46.21\%$.
\item \textbf{Scale yields diminishing returns under FT v4.} Across a $\sim 30\times$ parameter range, AvgPerf moves only $\sim 0.5$ points ($46.21{-}46.74$). The routing signal saturates well below 27B; capacity is not the bottleneck.
\item \textbf{Cost-aware regret requires per-task logs.} The archived larger-size sweep preserved aggregate AvgPerf but not per-task decisions, so we omit CumReg for $\geq 2$B sizes rather than report non-reproducible numbers. The available 0.8B run has CumReg $309.1$ under the same cost-aware reward $r=s-0.1\,\kappa$ used in Table~\ref{tab:main}.
\item \textbf{0.8B is a defensible cost choice.} Because larger sizes do not move the needle on AvgPerf, we report $0.8$B in Table~\ref{tab:main} as the most cost-efficient finetuned policy, and treat the larger sizes as scaling controls.
\end{itemize}

\section{Supplementary Analyses}
\label{app:analysis}

This section unpacks the in-distribution and OOD numbers in Table~\ref{tab:main} along several diagnostic axes: per-model capability profiles (\S\ref{app:radar}), the full $8\times 9$ score matrix on every split (\S\ref{app:full_matrix}), the structural information available to a router (\S\ref{app:variance}), the bias of a vanilla LLM router (\S\ref{app:bias_fig}), the dollar-cost regime decomposition (\S\ref{app:cost}), and the per-method breakdowns of both the in-distribution result (\S\ref{app:id_breakdown}) and the OOD result (\S\ref{app:ood_breakdown}). Unless explicitly noted for the updated GPT-5.4 OOD row, all numbers come from the canonical results bundle and are consistent with the headline figures in Table~\ref{tab:main}.

\subsection{Model Capability Profiles}
\label{app:radar}

\begin{figure}[h]
\centering
\includegraphics[width=0.6\textwidth]{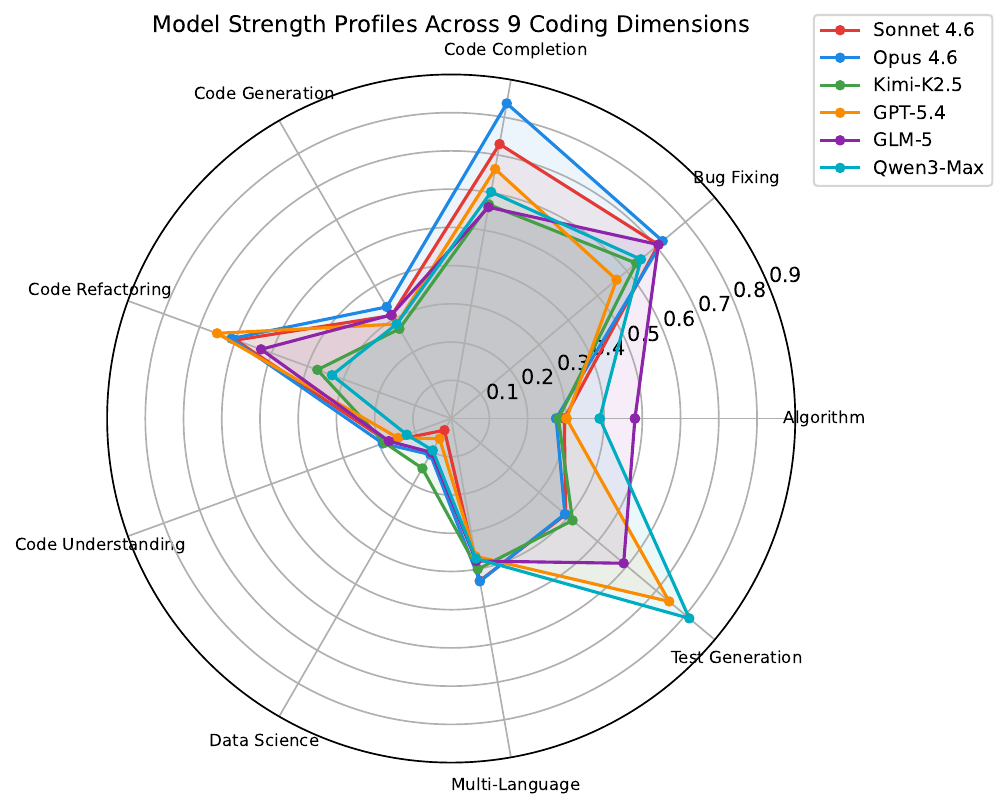}
\caption{Radar chart of representative models across 9 coding dimensions. Each model has a distinct strength profile, confirming the model complementarity that routing exploits. Axes are scaled per-dimension (0~=~worst model, 1~=~best model on that dimension).}
\label{fig:radar}
\end{figure}

To visualize how model strengths differ across the 9 single-turn coding dimensions, Figure~\ref{fig:radar} presents radar charts for representative models. Each axis represents one dimension; the radial distance is proportional to the model's average test-split score on that dimension (Table~\ref{tab:matrix}). The shapes are distinctly non-circular: Claude Opus excels on code completion and bug fixing but underperforms on algorithm design, while GLM-5 shows the reverse pattern. Qwen3-Max dominates test generation but is average elsewhere. These complementary profiles are precisely what routing exploits: no single model is best everywhere, and a dimension-aware router can select the right specialist for each task type.

\subsection{Full Model $\times$ Dimension Score Matrix}
\label{app:full_matrix}

Tables~\ref{tab:matrix}, \ref{tab:matrix_train}, and \ref{tab:matrix_full} present the model$\times$dimension matrix on the in-distribution test split, the probing (train) split, and the full single-turn corpus, respectively. Bold marks the probing-learned dimension-best model. Several patterns are visible: Claude~Opus~4.6 leads on code completion, code generation, and multi-language but ranks mid-pack on algorithm design; GLM-5 is the probing-learned dimension-best for algorithm design and bug fixing (test-split values $0.265$ and $0.607$, suppressed by GLM-5's $\sim$$14.5\%$ missing-task rate which contributes zeros under the canonical missing-as-$0$ convention); Qwen3-Max dominates test generation ($0.789$); Kimi-K2.5 leads on the two cheapest-to-call dimensions (data science and code understanding), where its low price tier yields the highest cost-adjusted upside.

\begin{table}[h]
\caption{Model $\times$ dimension score matrix on the \textbf{in-distribution test split} ($n{=}2{,}919$ across 9 single-turn dimensions, 60/10/30 md5 split). \textbf{Bold}~=~the dimension-best model learned on the probing split (Table~\ref{tab:matrix_train}). The bold (DimBest) cells are reported under the canonical missing-as-$0$ convention: any task with no recorded score for a model contributes $0$. The AVG column is the average performance of cross 9 dimensions; models with appreciable missing rates (GLM-5 $\sim$$14.5\%$, Qwen3.5-Plus $\sim$$10.7\%$) therefore have AVG below their scored-only mean.}
\label{tab:matrix}
\centering
\scriptsize
\resizebox{\textwidth}{!}{%
\begin{tabular}{lccccccccc|c}
\toprule
\textbf{Model} & \textbf{CdGen} & \textbf{Algo} & \textbf{Bug} & \textbf{Comp} & \textbf{Refac} & \textbf{DS} & \textbf{Multi} & \textbf{Und} & \textbf{TstGn} & \textbf{AVG} \\
\midrule
Claude Opus 4.6   & \textbf{.337} & .275          & .722          & \textbf{.837} & .612          & .109          & \textbf{.432} & .190          & .388          & \textbf{.438} \\
GPT-5.4           & .285          & .302          & .565          & .663          & \textbf{.652} & .061          & .366          & .148          & .744          & .422          \\
Claude Sonnet 4.6 & .311          & .296          & .707          & .729          & .600          & .035          & .431          & .177          & .391          & .413          \\
GLM-5             & .313          & \textbf{.265} & \textbf{.607} & .561          & .529          & .102          & .378          & .173          & .589          & .378          \\
Qwen3-Max         & .285          & .388          & .648          & .602          & .332          & .096          & .372          & .124          & \textbf{.789} & .400          \\
Qwen3.5-Plus      & .288          & .409          & .652          & .531          & .276          & .087          & .382          & .151          & .698          & .372          \\
Kimi-K2.5         & .271          & .280          & .632          & .568          & .372          & \textbf{.151} & .400          & \textbf{.185} & .415          & .367          \\
MiniMax-M2.7      & .245          & .069          & .504          & .583          & .610          & .125          & .361          & .179          & .471          & .361          \\
\bottomrule
\end{tabular}%
}
\end{table}

\begin{table}[h]
\caption{Model $\times$ dimension score matrix on the \textbf{probing set}. DimensionBest and trained classifiers use this split to learn per-dimension model rankings. Bold~=~best model per dimension on this split. AVG is the weighted average performance by per-dimension count.}
\label{tab:matrix_train}
\centering
\scriptsize
\resizebox{\textwidth}{!}{%
\begin{tabular}{lccccccccc|c}
\toprule
\textbf{Model} & \textbf{CdGen} & \textbf{Algo} & \textbf{Bug} & \textbf{Comp} & \textbf{Refac} & \textbf{DS} & \textbf{Multi} & \textbf{Und} & \textbf{TstGn} & \textbf{AVG} \\
\midrule
Claude Opus 4.6   & \textbf{.315} & .254          & .717          & \textbf{.860} & .607          & .142          & \textbf{.408} & .193          & .392          & \textbf{.429} \\
GPT-5.4           & .282          & .257          & .567          & .639          & \textbf{.644} & .063          & .346          & .150          & .764          & .412          \\
Claude Sonnet 4.6 & .275          & .258          & .698          & .751          & .615          & .068          & .407          & .180          & .395          & .402          \\
GLM-5             & .298          & \textbf{.472} & \textbf{.728} & .537          & .516          & .079          & .362          & .174          & .592          & .399          \\
Qwen3-Max         & .262          & .310          & .660          & .591          & .336          & .111          & .350          & .123          & \textbf{.827} & .392          \\
Qwen3.5-Plus      & .282          & .397          & .666          & .538          & .296          & .114          & .355          & .149          & .714          & .371          \\
Kimi-K2.5         & .269          & .254          & .653          & .590          & .386          & \textbf{.184} & .372          & \textbf{.195} & .430          & .370          \\
MiniMax-M2.7      & .239          & .073          & .528          & .563          & .603          & .145          & .331          & .184          & .494          & .360          \\
\bottomrule
\end{tabular}%
}
\end{table}

\begin{table}[h]
\caption{Model $\times$ dimension score matrix on the \textbf{full single-turn corpus} ($n{=}9{,}999$ across 9 single-turn dimensions, all splits combined). The AVG column is the pooled mean over all scored cells (i.e., weighted by per-dimension scored-task count), not the unweighted mean of 9 dimension means; for models with missing rates skewed toward stronger dimensions (GLM-5 $\approx 14.5\%$ missing, Qwen3.5-Plus $\approx 10.7\%$, MiniMax-M2.7 $\approx 5.4\%$, others $<2\%$) the pooled AVG runs slightly below the AVG. The OOD agentic-programming dimension is excluded.}
\label{tab:matrix_full}
\centering
\scriptsize
\resizebox{\textwidth}{!}{%
\begin{tabular}{lccccccccc|c}
\toprule
\textbf{Model} & \textbf{CdGen} & \textbf{Algo} & \textbf{Bug} & \textbf{Comp} & \textbf{Refac} & \textbf{DS} & \textbf{Multi} & \textbf{Und} & \textbf{TstGn} & \textbf{AVG} \\
\midrule
Claude Opus 4.6   & \textbf{.317} & .260          & \textbf{.719} & \textbf{.851} & .610          & .130          & \textbf{.414} & .194          & .396          & \textbf{.432} \\
Claude Sonnet 4.6 & .287          & .275          & .701          & .743          & .607          & .054          & .413          & .179          & .396          & .406          \\
GPT-5.4           & .280          & .270          & .565          & .643          & \textbf{.646} & .062          & .351          & .150          & .753          & .413          \\
GLM-5             & .297          & \textbf{.479} & .717          & .546          & .521          & .083          & .365          & .174          & .592          & .402          \\
Qwen3-Max         & .265          & .336          & .651          & .593          & .335          & .107          & .356          & .123          & \textbf{.823} & .394          \\
Qwen3.5-Plus      & .282          & .405          & .661          & .533          & .292          & .105          & .362          & .150          & .706          & .371          \\
Kimi-K2.5         & .268          & .265          & .645          & .584          & .380          & \textbf{.173} & .378          & \textbf{.194} & .423          & .369          \\
MiniMax-M2.7      & .240          & .070          & .517          & .572          & .603          & .134          & .338          & .183          & .480          & .358          \\
\bottomrule
\end{tabular}%
}
\end{table}

\subsection{Variance Decomposition}
\label{app:variance}

To quantify how much of the per-task oracle decision is recoverable from cheap structural signals, we compute the mutual information $I(y_t;\, d(t))$ between the oracle-assigned model $y_t = \arg\max_m s(t, m)$ and the task dimension $d(t)$ on the in-distribution test split. We find that dimension identity captures roughly 27\% of the entropy of $y_t$: enough to explain why the DimensionBest reaches about $0.475$ AvgPerf, but well short of the $0.570$ per-task oracle. The remainder of the routing signal lies in the per-task content (algorithm choice, API patterns, edge-case handling), which is exactly what ACRouter's task-embedding Memory keys on rather than the lower-resolution dimension hash used by DimensionBest.

\subsection{Router Bias Distribution}
\label{app:bias_fig}

\begin{figure}[h]
\centering
\includegraphics[width=0.85\textwidth]{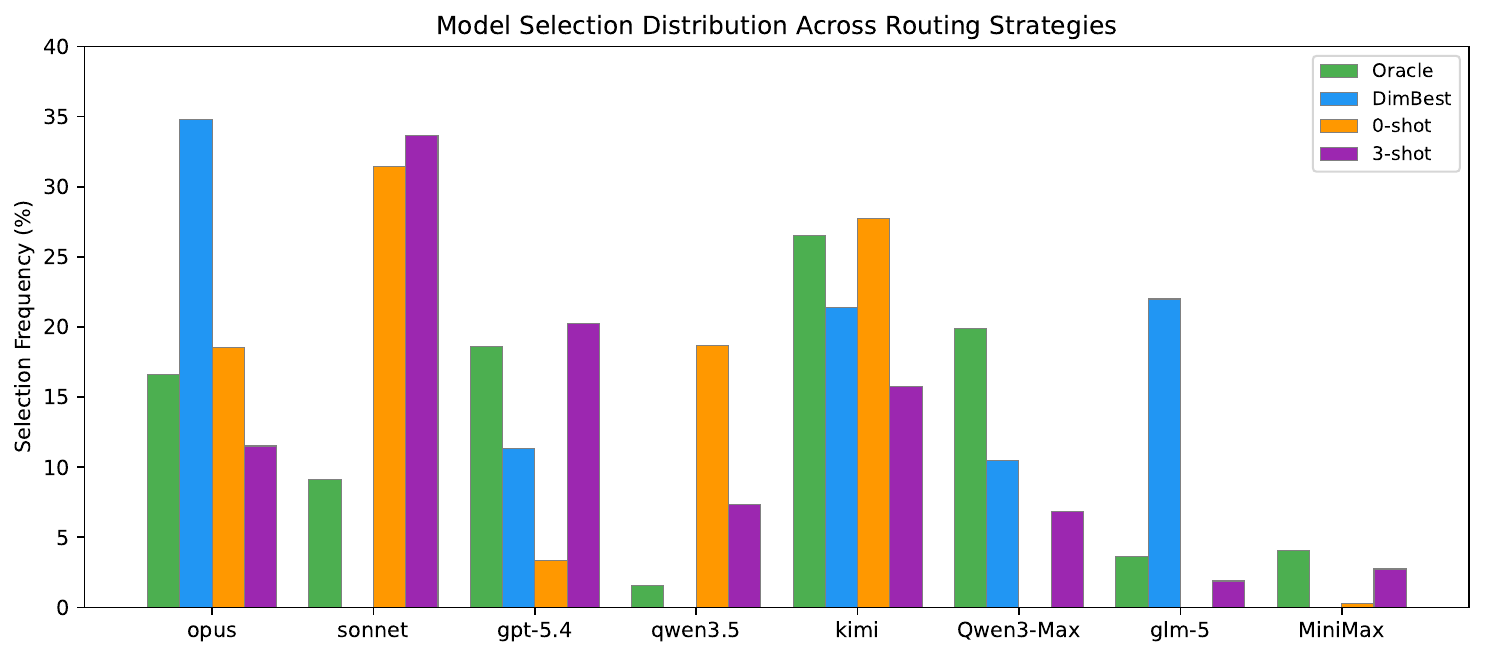}
\caption{Router bias analysis: model-selection distributions of the Vanilla LLM router (left) versus the per-task Oracle (right) across the 9 single-turn dimensions. The Vanilla router's selections are nearly uniform across dimensions, failing to exploit the dimensional structure.}
\label{fig:bias}
\end{figure}

Figure~\ref{fig:bias} shows the model-selection distributions of the LLM router (Vanilla in Table~\ref{tab:bias}) versus the per-task Oracle across dimensions. The Vanilla router's selections are nearly uniform, failing to exploit the strong dimensional structure that DimensionBest leverages. This visual diagnosis is consistent with the +Perf~stats finding in Table~\ref{tab:bias}: providing the per-dimension performance prior lets the LLM router match (and slightly exceed) DimensionBest with no architectural change, confirming that the bottleneck is information rather than reasoning.

\subsection{Cost Comparison}
\label{app:cost}

This subsection decomposes the dollar-cost columns of Table~\ref{tab:main} into three deployment-cost regimes. Costs are computed using the per-model asymmetric input/output pricing in Table~\ref{tab:pricing} and include router-side overhead where applicable (priced at \$0.054/M for self-hosted policies, see \S\ref{app:local_price}).

Three regimes emerge consistent with Table~\ref{tab:main}. The \emph{cheap} tier (trained classifiers \$7--8: LogReg \$7.54, RouteLLM-MF \$7.46, TF-IDF$+$MLP \$7.69, RouteLLM-BERT \$7.59, Qwen3.5-0.8B-Finetuned \$6.81) attains near-DimensionBest performance at minimal cost: these methods add little router token overhead and tend to route a substantial share of tasks to mid-tier models (Kimi-K2.5, Qwen3-Max). The \emph{moderate} tier (DimensionBest \$12.89, ACRouter \$13.21, online bandits \$10--11) pays for an informed mix of expensive and cheap backend models in exchange for higher per-task quality; ACRouter additionally pays for its kNN Memory and Verifier overhead. The \emph{premium} tier (Always-Opus \$34.02, Always-Qwen3.5+ \$18.14, Random \$15.64) commits to a costly default. Always-Opus is not cost-efficient on ID: it spends \$34.02 for $43.83\%$ AvgPerf, while ACRouter achieves $49.98\%$ AvgPerf for \$13.21. The cheapest single-model option, Always-Kimi-K2.5 at \$2.90, sets the cost floor with $36.66\%$ AvgPerf (Perf/\$ $12.62$, the highest in Table~\ref{tab:main}), at a substantial AvgPerf cost compared with ACRouter.

\subsection{In-distribution Per-Method Breakdown}
\label{app:id_breakdown}

Table~\ref{tab:id_breakdown} reports the per-method breakdown on the in-distribution test ($n{=}2{,}919$ across 9 single-turn coding dimensions) corresponding to the left-side block of Table~\ref{tab:main}, with the dollar-cost column made explicit. AvgPerf\% is the average performance of execution-graded scores across dimensions, and CumReg uses the per-task cost-aware reward $r=s-0.1\,\kappa_{\mathrm{USD}}$ against the per-task oracle.

\begin{table}[h]
\caption{Per-method breakdown on the in-distribution test (2{,}919 tasks across 9 single-turn coding dimensions). Costs include both backend and router-side tokens (self-hosted at \$0.054/M, API at Table~\ref{tab:pricing} rates). CumReg uses the cost-aware reward $r=s-0.1\,\kappa_{\mathrm{USD}}$.}
\label{tab:id_breakdown}
\centering
\small
\resizebox{\textwidth}{!}{%
\begin{tabular}{llcccc}
\toprule
\textbf{Family} & \textbf{Router}                & \textbf{AvgPerf\%}$\uparrow$ & \textbf{CumReg}$\downarrow$ & \textbf{\$Total} & \textbf{Perf/\$}$\uparrow$ \\
\midrule
Bound          & Oracle (cost-aware)             & 57.00          & 0              & \$6.95   & 8.20          \\
\midrule
Agentic        & ACRouter (ours)                 & \textbf{49.98} & \textbf{205.5} & \$13.21  & 3.79          \\
\midrule
Bandit         & LinTS (5-seed avg)              & 46.48          & 307.4          & \$10.35  & 4.49          \\
Bandit         & LinUCB (5-seed avg)             & 46.84          & 296.9          & \$10.69  & 4.38          \\
\midrule
Heuristic      & DimensionBest                   & 47.50          & 277.4          & \$12.89  & 3.69          \\
Heuristic      & kNN Retrieval                   & 47.18          & 286.7          & \$7.77   & 6.07          \\
\midrule
Trained        & LogReg                          & 47.26          & 284.4          & \$7.54   & 6.27          \\
Trained        & RouteLLM-BERT                        & 47.22          & 285.5          & \$7.59   & 6.22          \\
Trained        & TF-IDF$+$MLP                    & 46.97          & 292.8          & \$7.69   & 6.11          \\
Trained        & Qwen3.5-0.8B-Finetuned          & 46.41          & 309.1          & \$6.81   & 6.82          \\
Trained        & RouteLLM-MF                     & 46.16          & 316.5          & \$7.46   & 6.19          \\
\midrule
Single-Model   & Always-Opus 4.6                 & 43.83          & 387.1          & \$34.02  & 1.29          \\
Single-Model   & Always-Kimi-K2.5                & 36.66          & 593.3          & \$2.90   & \textbf{12.62}\\
Single-Model   & Always-Qwen3.5-Plus             & 37.16          & 580.2          & \$18.14  & 2.05          \\
Single-Model   & Random                          & 38.75          & 533.6          & \$15.64  & 2.48          \\
\bottomrule
\end{tabular}%
}
\end{table}

Three patterns mirror the OOD breakdown but with sharper separation. ACRouter's $49.98\%$ AvgPerf and $205.5$ CumReg dominate every other family on this in-distribution split, beating DimensionBest's full dimension-level prior by $2.48$ AvgPerf points at $\$13.21$ vs.\ $\$12.89$. Trained classifiers cluster near DimensionBest in AvgPerf ($46.16{-}47.26\%$) but at roughly half the cost ($\$6.81{-}\$7.69$), giving them the strongest Perf/\$ among non-oracle methods that exceed $46\%$ AvgPerf. Always-Opus pays the most ($\$34.02$) yet trails ACRouter by $6.15$ AvgPerf points and accumulates $387.1$ regret, illustrating that committing to the strongest single model is not cost-efficient even on the controlled split.

\subsection{Real-World Test (OOD) Per-Method Breakdown}
\label{app:ood_breakdown}

Table~\ref{tab:ood_breakdown} reports the per-method breakdown on the real-world OOD test (176 agentic-programming tasks) corresponding to the right-side block of Table~\ref{tab:main}. AvgPerf\% is the resolved-rate ($\#$resolved $/$ 176). CumReg uses the same per-task reward $r=s-0.1\,\kappa_{\mathrm{USD}}$ as the ID evaluation. The OOD setting is intentionally distribution-shifted from the 9 single-turn dimensions, and amplifies the difference between routers that rely on a frozen probing-set prior (which transfers poorly) and routers that adapt online (Memory + Verifier).

\begin{table}[h]
\caption{Per-method breakdown on the OOD test (176 SWE-bench-derived tasks). Routers without OOD coverage (e.g., DimensionBest, since the agentic dimension is held out) are marked ``---''. Costs include both backend and router-side tokens (self-hosted at \$0.054/M, API at Table~\ref{tab:pricing} rates).}
\label{tab:ood_breakdown}
\centering
\small
\resizebox{\textwidth}{!}{%
\begin{tabular}{llcccc}
\toprule
\textbf{Family} & \textbf{Router}                & \textbf{AvgPerf\%}$\uparrow$ & \textbf{CumReg}$\downarrow$ & \textbf{\$Total} & \textbf{Perf/\$}$\uparrow$ \\
\midrule
Bound          & Oracle (cost-aware)             & 75.89          & 0    & \$32.71  & 2.32          \\
\midrule
Agentic        & ACRouter (ours)                 & \textbf{62.50} & \textbf{17.0} & \$52.97  & 1.18          \\
\midrule
Bandit         & LinTS (5-seed avg)              & 46.43          & 35.9 & \$61.91  & 0.75          \\
Bandit         & LinUCB (5-seed avg)             & 49.82          & 31.1 & \$51.90  & 0.96          \\
\midrule
Heuristic      & DimensionBest                   & ---            & ---  & ---      & ---           \\
Heuristic      & kNN Retrieval                   & 14.29          & 66.7 & \$9.86   & \textbf{1.45}          \\
\midrule
Trained        & LogReg                          & 19.64          & 61.8 & \$16.79  & 1.17          \\
Trained        & RouteLLM-BERT                        & 21.43          & 59.4 & \$16.48  & 1.30 \\
Trained        & TF-IDF$+$MLP                    & 13.39          & 67.9 & \$11.44  & 1.17          \\
Trained        & Qwen3.5-0.8B-Finetuned          & 55.36          & 27.2 & \$74.81  & 0.74          \\
Trained        & RouteLLM-MF                     &  8.93          & 72.7 & \$9.50   & 0.94          \\
\midrule
Single-Model   & Always-Opus 4.6                 & 57.14          & 26.7 & \$89.28  & 0.64          \\
Single-Model   & Always-Kimi-K2.5                & 18.75          & 62.3 & \$15.37  & 1.22          \\
Single-Model   & Always-Qwen3.5-Plus             &  2.68          & 80.1 & \$14.11  & 0.19          \\
Single-Model   & Random                          & 31.25          & 50.4 & \$36.76  & 0.85          \\
\bottomrule
\end{tabular}%
}
\end{table}

\paragraph{Why the static baselines collapse on OOD.}
The 9 single-turn probing dimensions are dominated by short, single-file tasks; the OOD set requires multi-step planning, file navigation, and iterative debugging. Trained classifiers (LogReg, TF-IDF$+$MLP, RouteLLM variants) condition on prompt features that no longer carry the same signal in the OOD distribution, and they cannot acquire new feedback during evaluation. ACRouter and the online bandits, by contrast, accumulate Memory or per-arm posteriors during the OOD stream and recover more signal. ACRouter's stronger Orchestrator + Memory routing reaches $62.50\%$ resolved-rate compared with $46{-}50\%$ for the bandits, and its $17.0$ cost-aware CumReg ranks first across the table, which is ahead of Always-Opus ($26.7$).

\paragraph{Honest tracking of agentic outcomes.}
The OOD AvgPerf column is the harness-graded resolved-rate ($\#$resolved $/$ 176) and not a looser \texttt{apply\_ok} signal (which only checks that the submitted patch applies cleanly without verifying that the repository's tests pass).

\paragraph{Per-Model Score Matrix on OOD.}
Table~\ref{tab:ood_matrix} reports each candidate backend's headline metrics on the same 176-task OOD set, analogous to Table~\ref{tab:matrix} on the in-distribution split. Because the OOD set spans a single dimension of agentic programming, the matrix degenerates to a per-model column with per-model average call/cost statistics added for diagnostic transparency. Unlike the in-distribution matrix where five different models serve as dimension-best, OOD performance is more strongly ordered by base coding capability: the updated GPT-5.4 run resolves $132/176 = 75.00\%$ of tasks, Opus reaches $57.14\%$, Sonnet reaches $49.11\%$, and the long tail drops below $30\%$. The per-task oracle row is retained from the original 8-backend score bundle and resolves $75.89\%$ of tasks.

\begin{table}[h]
\caption{Per-model score matrix on the 176-task OOD set. \textbf{Resolved\%}: fraction of tasks where the model produces a patch that the SWE-Bench harness grades as correct. \textbf{Apply\_ok\%}: fraction whose patch at least applies cleanly to the test repo, regardless of test outcome. \textbf{Calls/task}: average number of backend invocations per task across all retries permitted by the harness (capped at 40 steps). \textbf{\$Total}: total backend dollar cost over all 176 tasks under the pricing in Table~\ref{tab:pricing}. The GPT-5.4 row is updated from the latest OOD run on the same split; Apply\_ok\% and Calls/task are not logged for that run. \textbf{Per-task oracle}: retained from the original 8-backend score bundle.}
\label{tab:ood_matrix}
\centering
\small
\resizebox{\textwidth}{!}{%
\begin{tabular}{lcccc}
\toprule
\textbf{Model} & \textbf{Resolved\%}$\uparrow$ & \textbf{Apply\_ok\%}$\uparrow$ & \textbf{Calls/task} & \textbf{\$Total} \\
\midrule
Claude Opus 4.6     & \textbf{57.14} & 79.46          & 19.0 & \$89.28 \\
Claude Sonnet 4.6   & 49.11          & 68.75          & 16.4 & \$77.23 \\
GPT-5.4             & 75.00          & ---            & ---  & \$41.04 \\
GLM-5               & 28.57          & 41.07          & 15.3 & \$26.73 \\
Kimi-K2.5           & 18.75          & 26.79          & 12.3 & \$15.35 \\
MiniMax-M2.7        & 14.29          & 17.86          &  7.1 & \$3.58  \\
Qwen3-Max           &  8.93          & 19.64          &  5.9 & \$9.58  \\
Qwen3.5-Plus        &  2.68          &  5.36          & 15.4 & \$14.38 \\
\midrule
\textbf{Per-task oracle} & \textbf{75.89} & --- & --- & \$32.71 \\
\bottomrule
\end{tabular}%
}
\end{table}

\section{Discussion}
\label{sec:discussion}

\subsection{Why Routing Matters}
\label{app:disc_persists}

Model complementarity is not a transient artifact. Each new model generation introduces new strengths (GLM-5 on algorithms, Qwen3-Max on test generation, Kimi-K2.5 on data science). Cost differentials persist: the most expensive model in our pool (Claude Opus 4.6 output at \$25/M) is over $20\times$ the cheapest (MiniMax-M2.7 output at \$1.20/M, Table~\ref{tab:pricing}). As the ecosystem expands with open-source, domain-specialized, and reasoning-specialized models, routing value increases. Routing is a permanent component of agent systems.

\subsection{Practitioner's Guide: Building Your Own Agent Router}
\label{app:disc_guide}

The core output of this work is not a leaderboard but a construction kit:
\begin{enumerate}[nosep,leftmargin=*]
\item \textbf{Set up the tool layer}: Plug in your candidate models and configure the execution sandbox.
\item \textbf{Profile via probing set}: Use CodeRouterBench (or your own tasks via C-A-F) to build a dimension$\times$model performance matrix.
\item \textbf{Start with DimensionBest}: A static Memory + lookup achieves about 83\% of oracle AvgPerf at near-zero overhead. This is your baseline.
\item \textbf{Add a classifier}: Swap the routing tool to a trained classifier (LogReg or RouteLLM) for cheaper deployment with comparable AvgPerf to DimensionBest. Trained classifiers achieve Perf/\$ of 6.11--6.82 in Table~\ref{tab:main}.
\item \textbf{Complete the C-A-F loop (ACRouter)}: When deploying on new distributions, activate all three modules to close the feedback loop. Initialize Memory with whatever priors you have; the C-A-F loop ensures convergence.
\item \textbf{Customize tools}: Swap evaluation tools in the Verifier (e.g., domain-specific tests), add custom routing tools.
\item \textbf{Extend the benchmark}: Add tasks via C-A-F. New models need responses + scoring; new dimensions need a task set + scoring function.
\end{enumerate}

\subsection{Beyond Model Routing}
\label{app:disc_beyond}

The C-A-F loop (observe context, act, receive feedback, update context) is not specific to model routing. The same paradigm applies to tool selection, API endpoint selection, prompt strategy selection, and thinking-effort allocation. We view coding routing as a concrete first instantiation of a broader agent decision-making paradigm. Future work can extend this to skill routing, sub-agent routing, memory routing, and effort routing.

\subsection{Extensibility Roadmap}
\label{app:disc_roadmap}

V1 (this release) provides 10 dimensions, 8 models, $\sim 80$K execution-verified responses. In V2, new models join by generating responses on the existing task set ($\sim 11$K API calls per model). In V3, new dimensions join by providing C-A-F triples with a scoring function. The long-term goal is a community-maintained living protocol where researchers contribute dimensions, models, and routing methods through standardized interfaces.

\section{Task Examples Across All 10 Dimensions}
\label{app:cases}

\begin{figure}[h]\centering
\includegraphics[width=\textwidth]{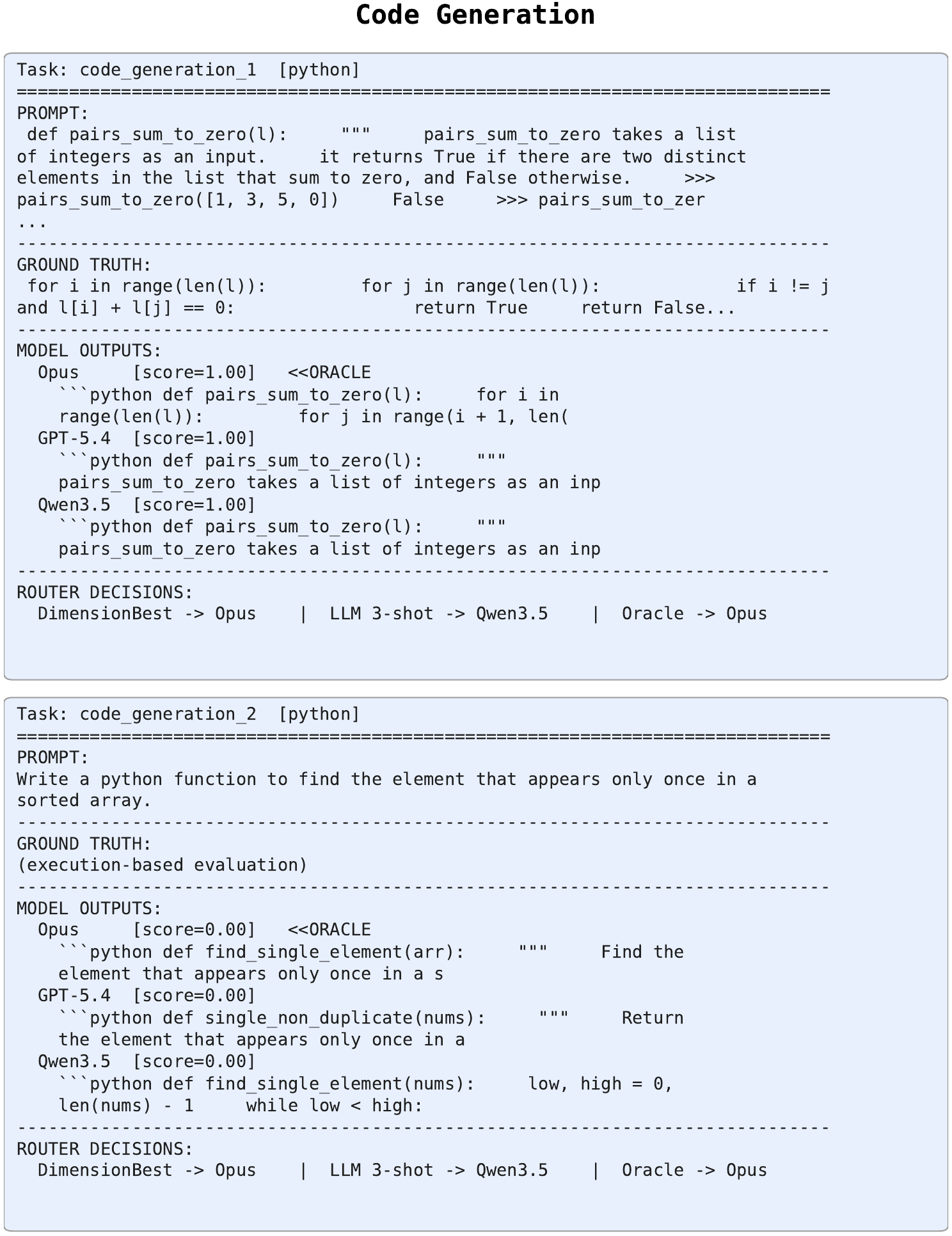}
\caption{Code Generation examples.}\label{fig:case_codegen}
\end{figure}

\begin{figure}[h]\centering
\includegraphics[width=\textwidth]{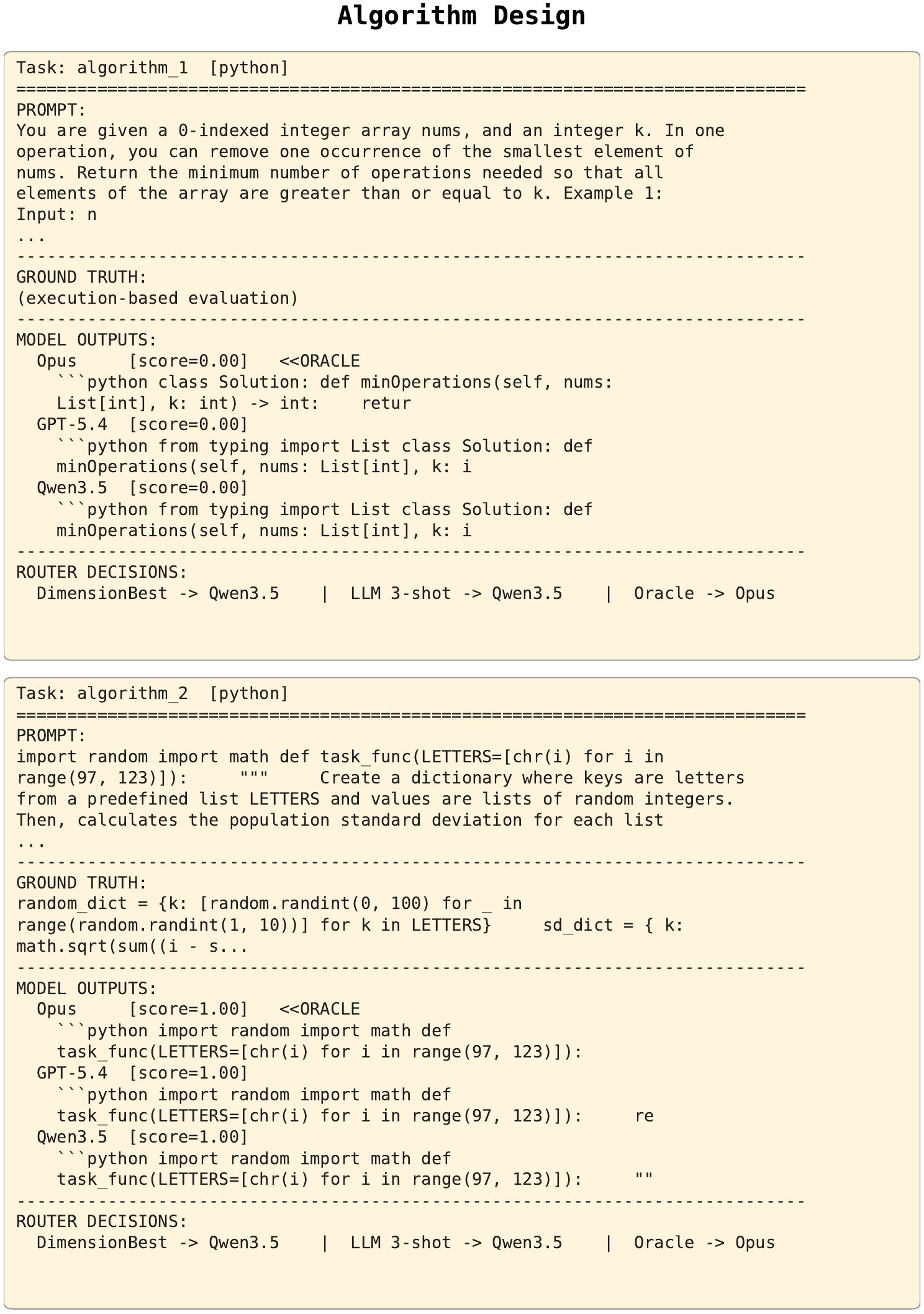}
\caption{Algorithm Design examples.}\label{fig:case_algo}
\end{figure}

\begin{figure}[h]\centering
\includegraphics[width=\textwidth]{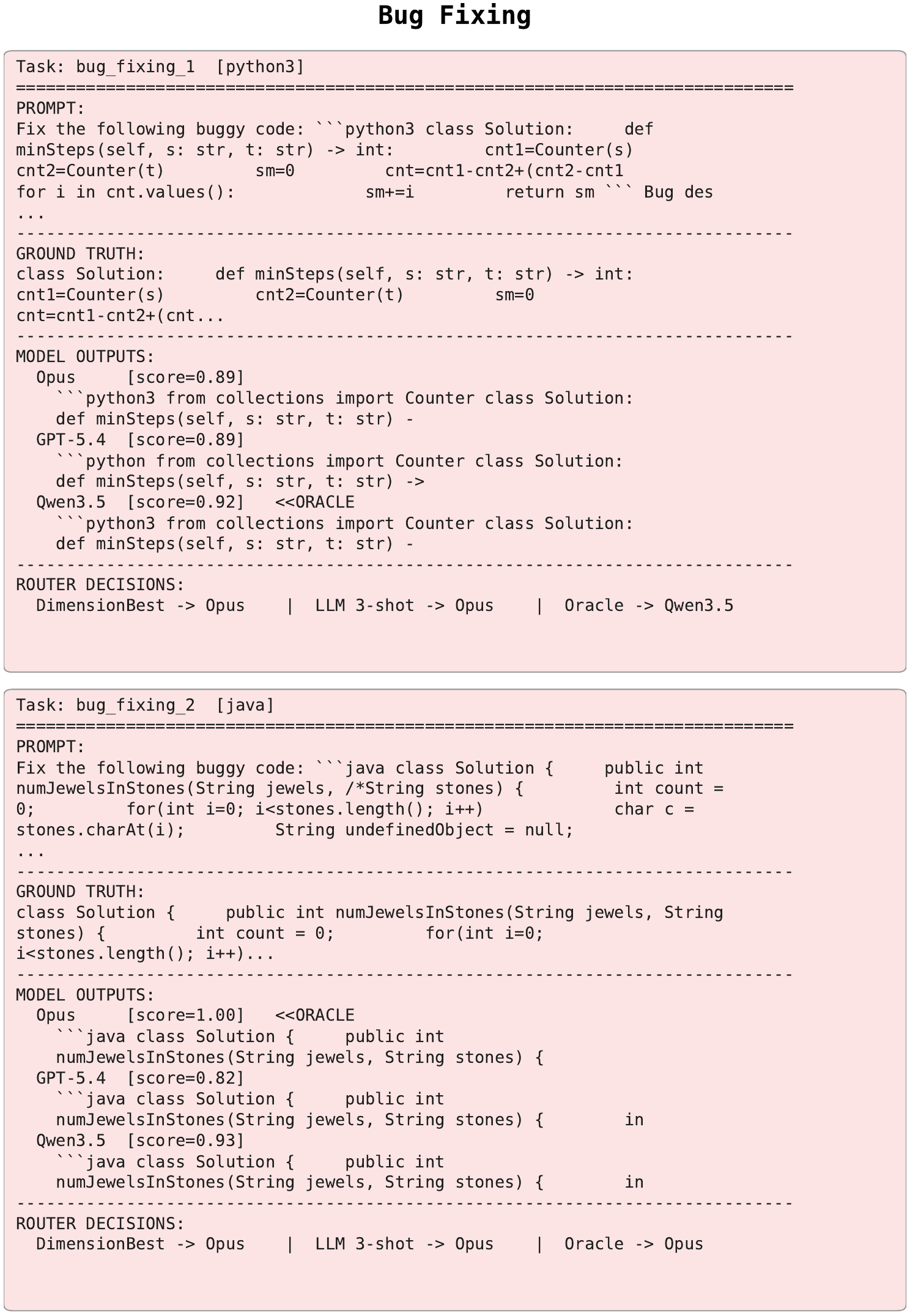}
\caption{Bug Fixing examples.}\label{fig:case_bugfix}
\end{figure}

\begin{figure}[h]\centering
\includegraphics[width=\textwidth]{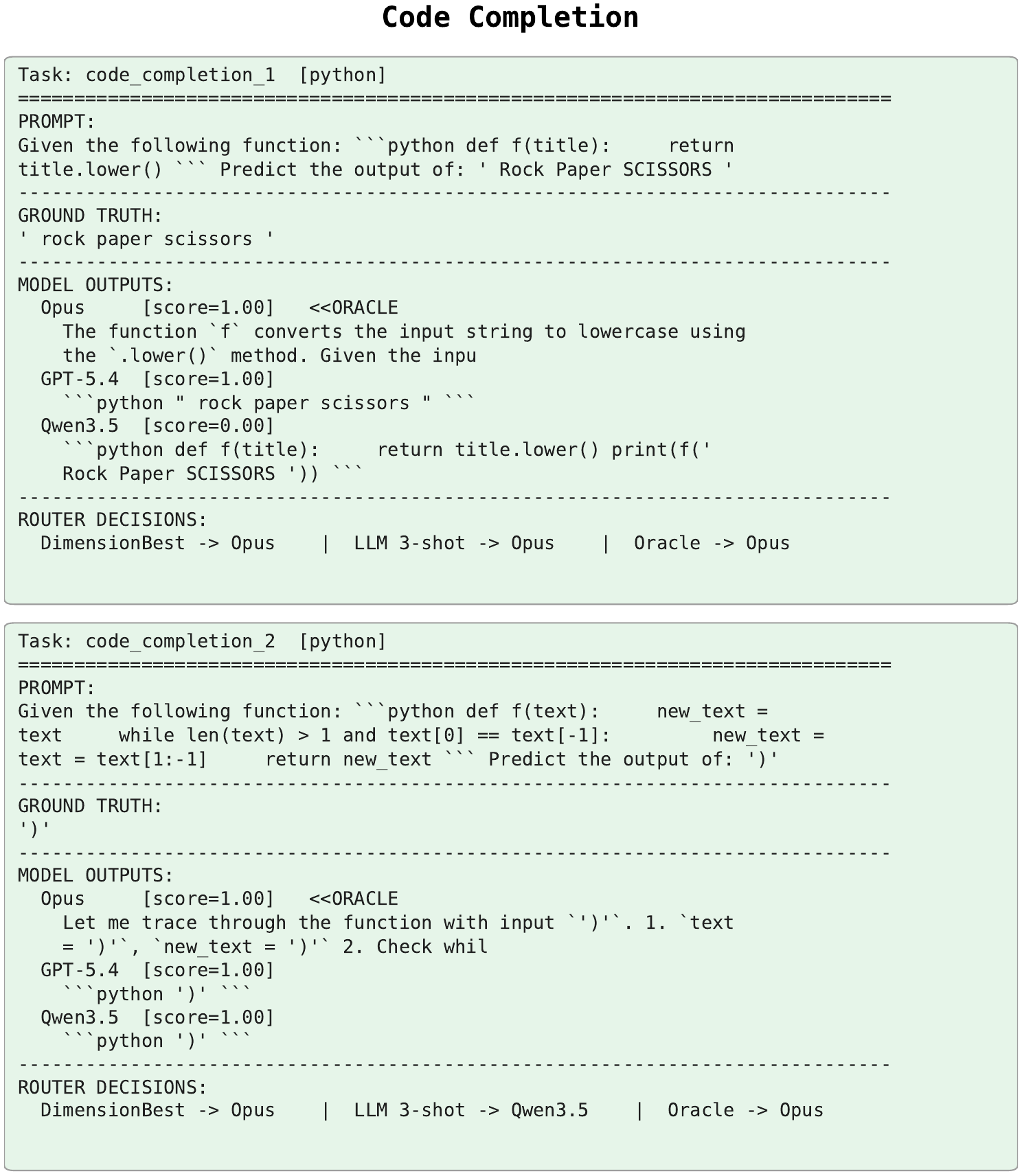}
\caption{Code Completion examples.}\label{fig:case_comp}
\end{figure}

\begin{figure}[h]\centering
\includegraphics[width=\textwidth]{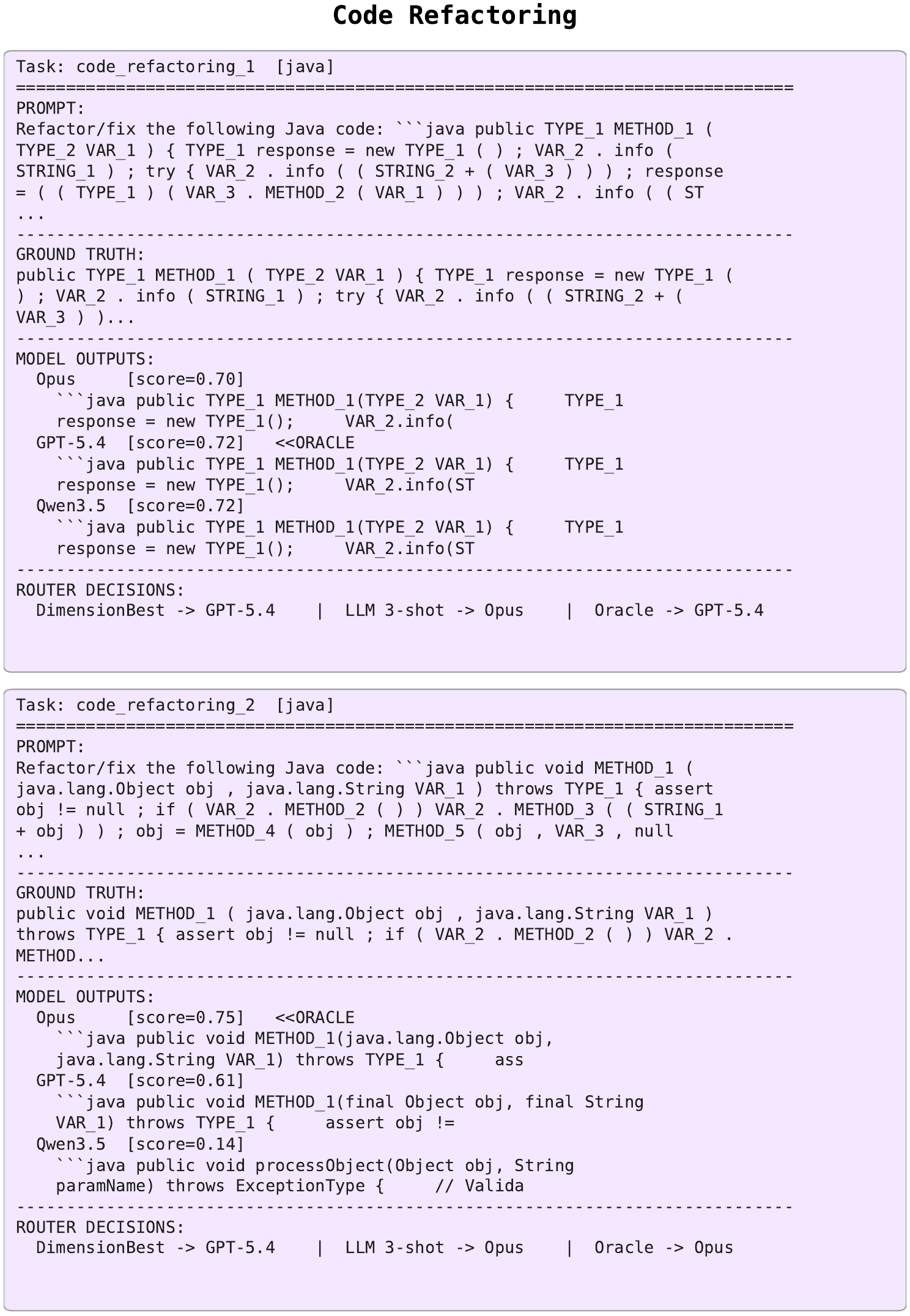}
\caption{Code Refactoring examples.}\label{fig:case_refac}
\end{figure}

\begin{figure}[h]\centering
\includegraphics[width=\textwidth]{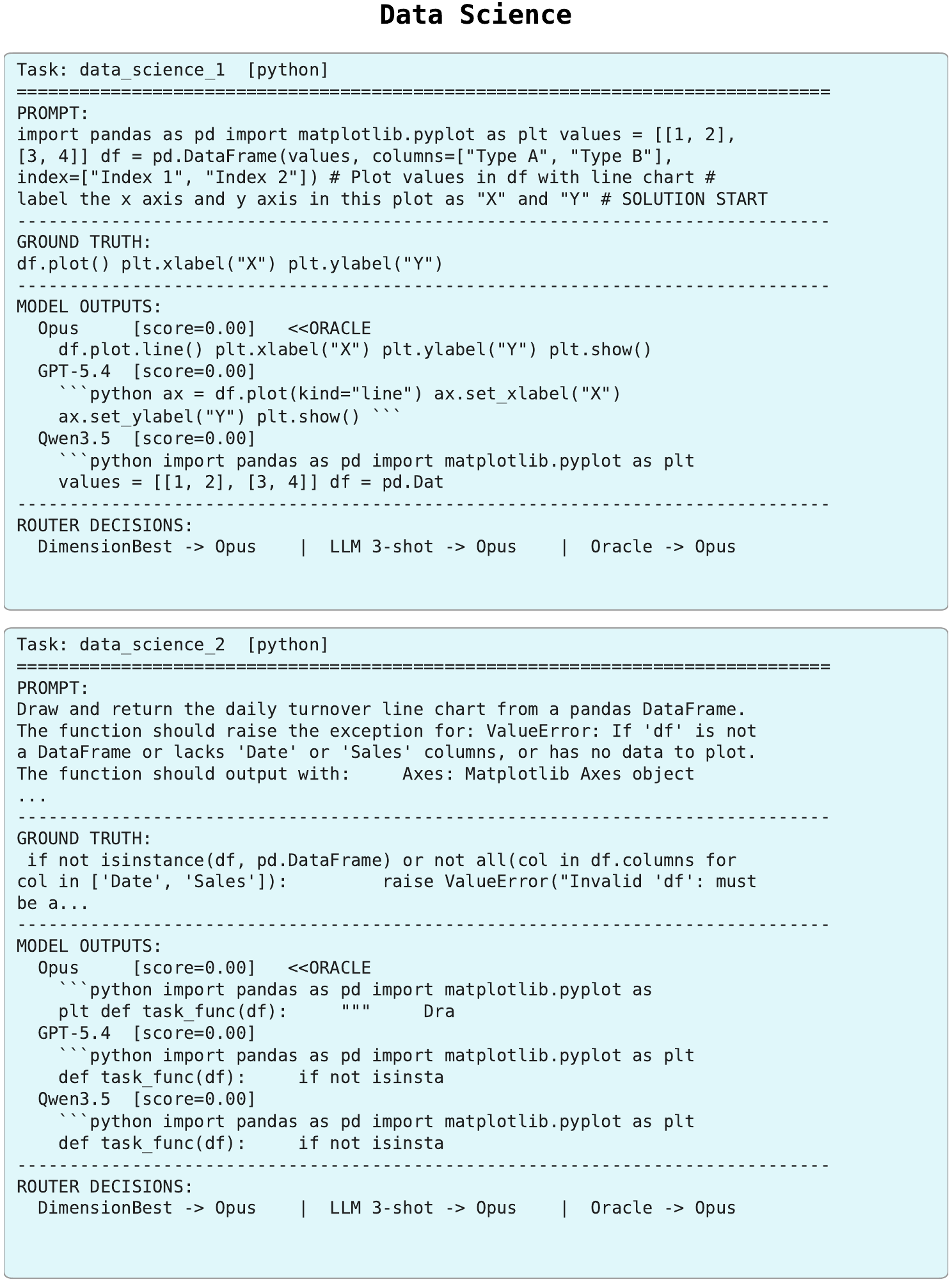}
\caption{Data Science examples.}\label{fig:case_ds}
\end{figure}

\begin{figure}[h]\centering
\includegraphics[width=\textwidth]{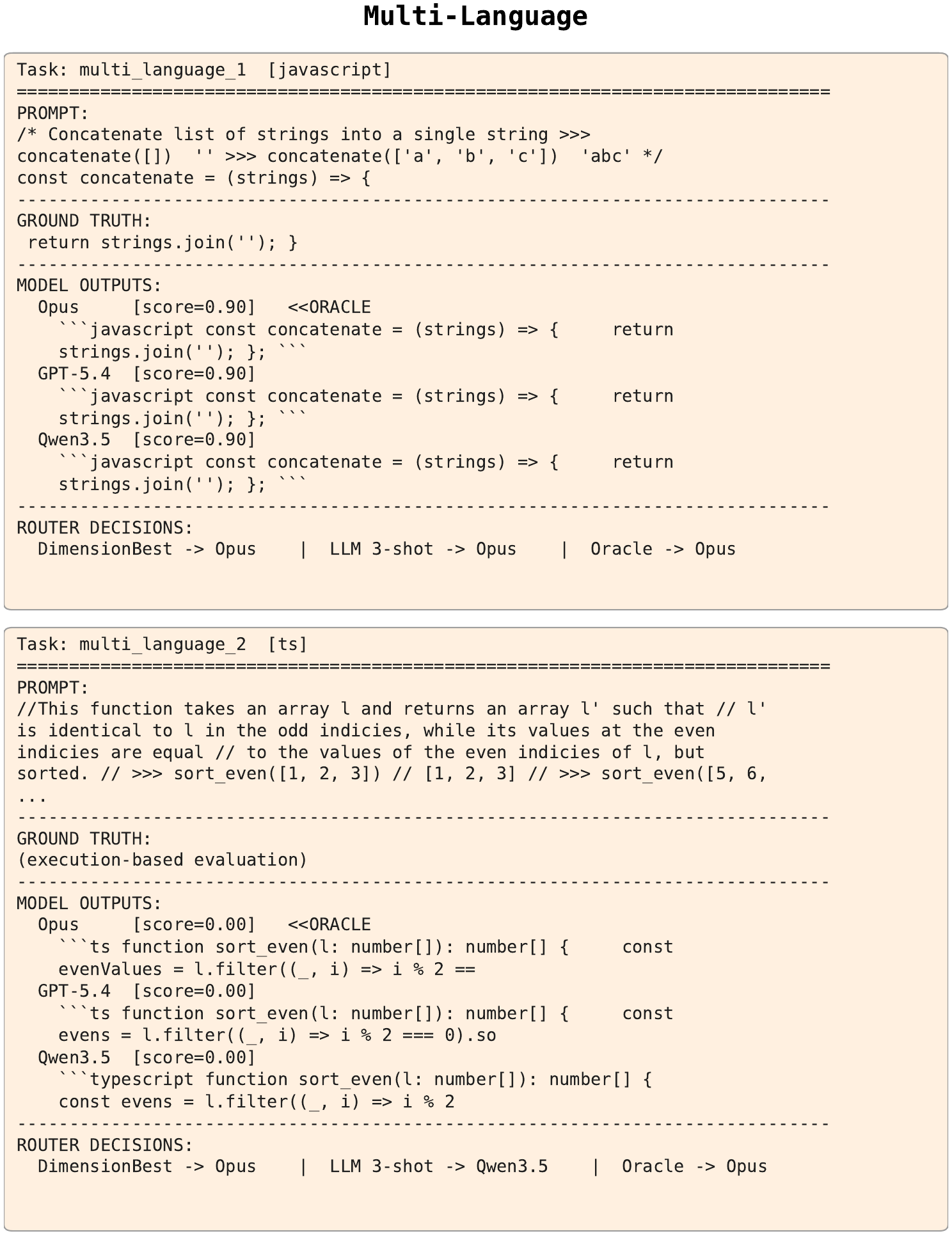}
\caption{Multi-Language examples.}\label{fig:case_multi}
\end{figure}

\begin{figure}[h]\centering
\includegraphics[width=\textwidth]{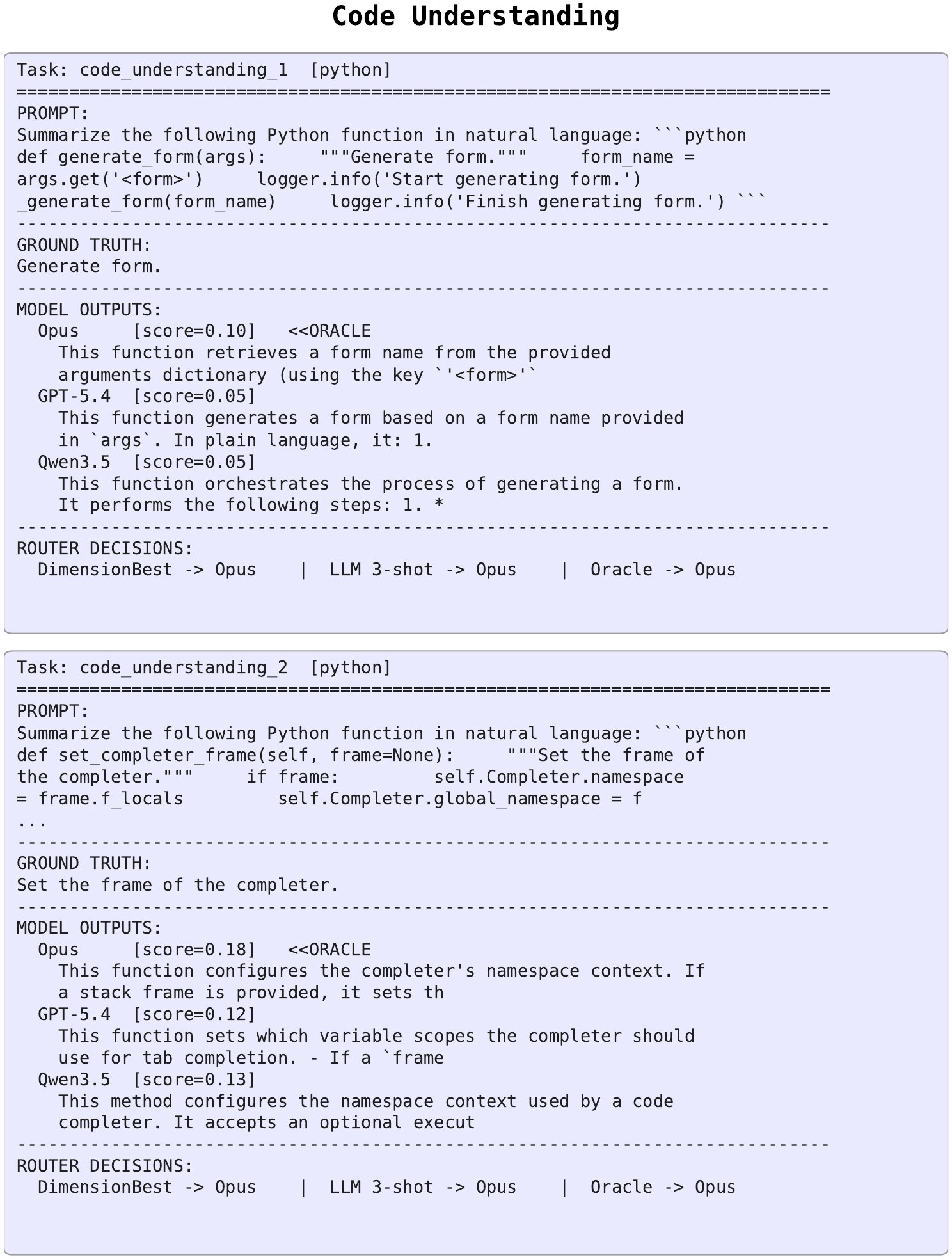}
\caption{Code Understanding examples.}\label{fig:case_und}
\end{figure}

\begin{figure}[h]\centering
\includegraphics[width=\textwidth]{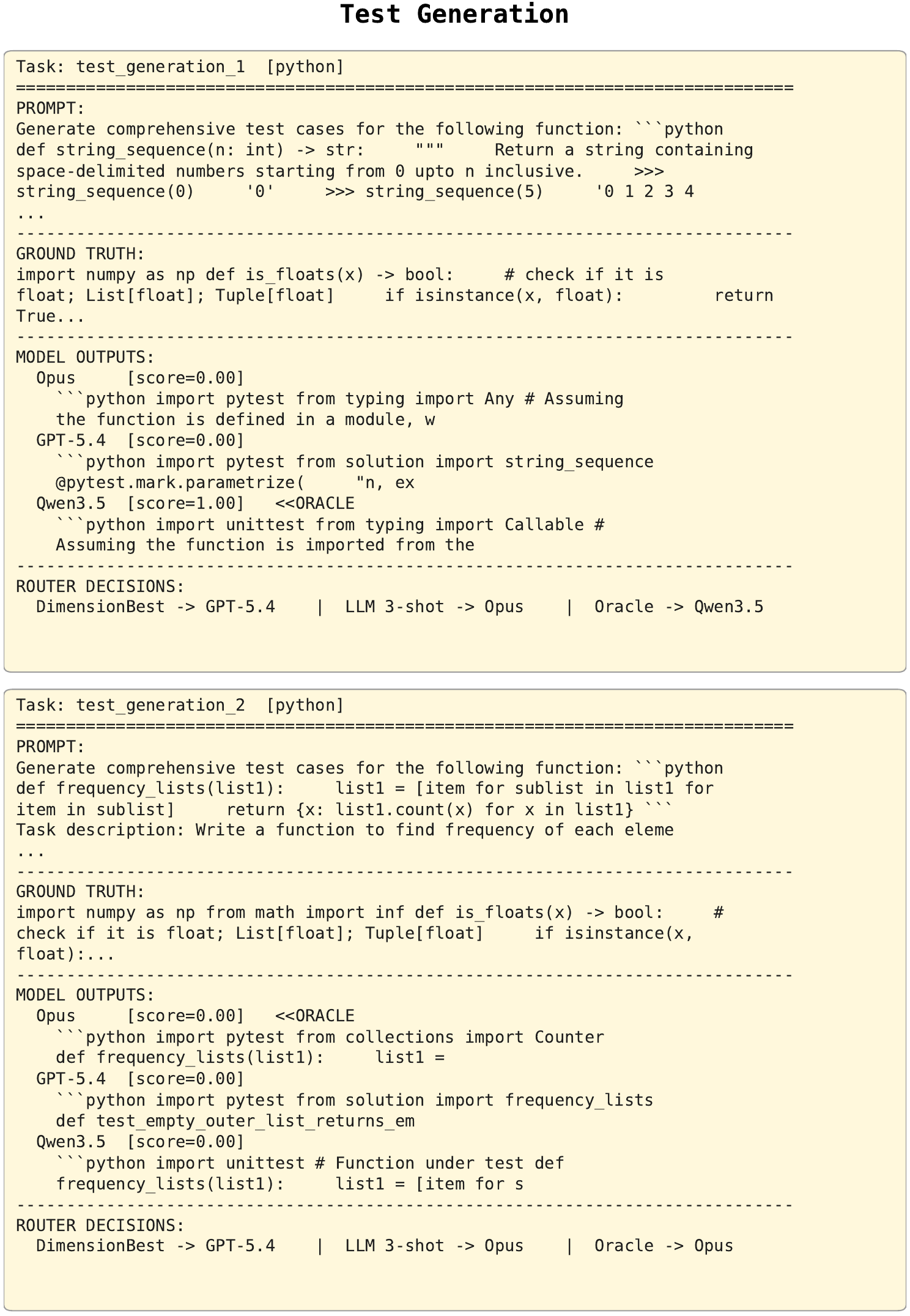}
\caption{Test Generation examples.}\label{fig:case_testgen}
\end{figure}

\begin{figure}[h]\centering
\includegraphics[width=\textwidth]{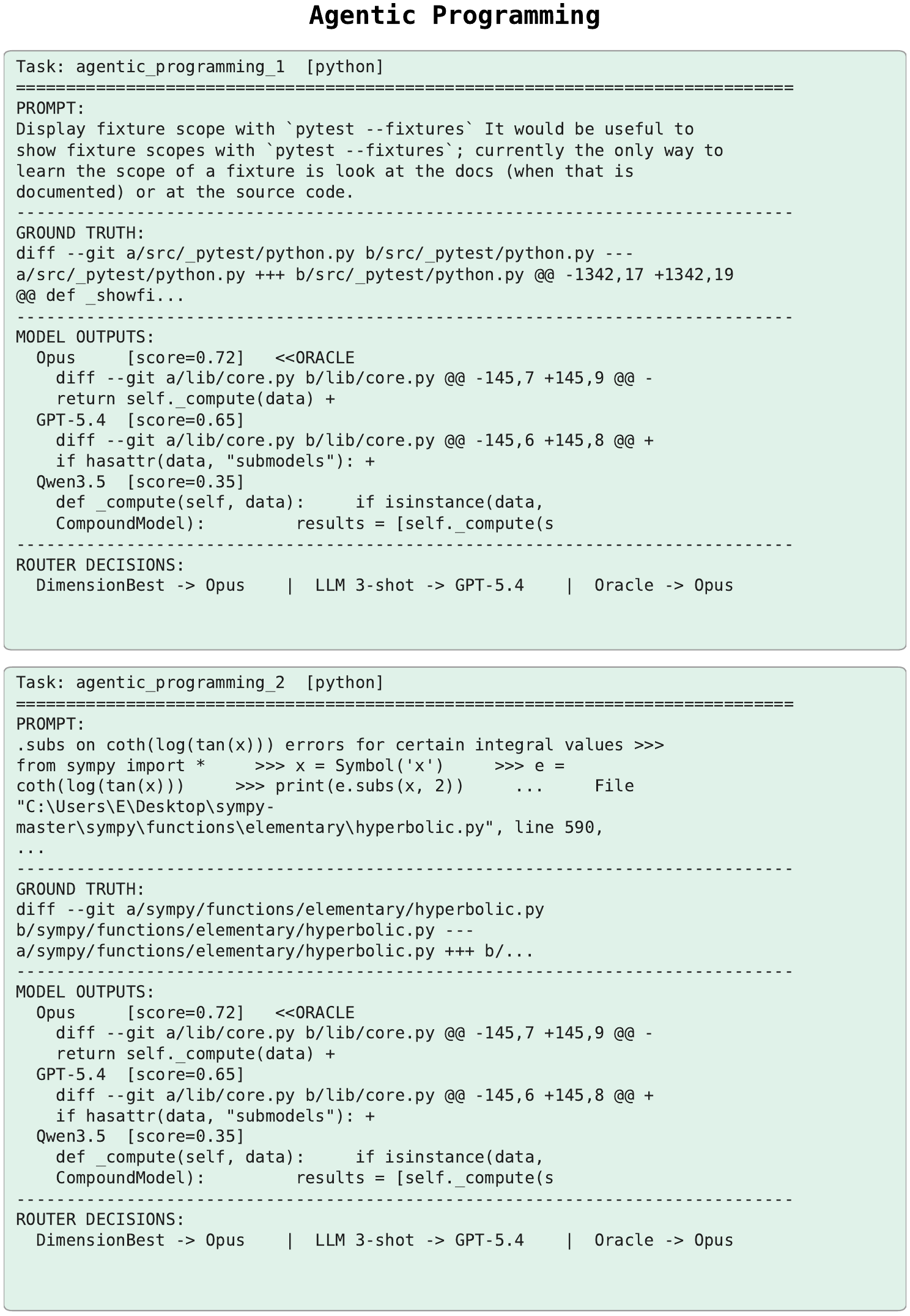}
\caption{Agentic Programming examples (10th dimension, OOD).}\label{fig:case_agentic}
\end{figure}

Figures~\ref{fig:case_codegen}--\ref{fig:case_agentic} present two representative task examples from each of the 10 dimensions, showing the input prompt, the best model's response, and per-model scores. These examples illustrate the diversity of coding tasks in CodeRouterBench and the model complementarity that routing exploits.


\clearpage

\end{document}